\documentclass[3p, final]{elsarticle}

\makeatletter
\def\ps@pprintTitle{%
   \let\@oddhead\@empty
   \let\@evenhead\@empty
   \let\@oddfoot\@empty
   \let\@evenfoot\@oddfoot
}
\makeatother

\usepackage{amssymb}

\usepackage{amsmath,booktabs,bm,url}
\usepackage[dvipdfmx]{color}

\newcommand{\ds}{\displaystyle}

\begin{document}

\begin{frontmatter}



\title{A Latent-class Model for Estimating Product-choice Probabilities from Clickstream Data}


\author[label1]{Naoki Nishimura}
\address[label1]{Product Management Unit, Internet Business Development Division, \\
Recruit Lifestyle Co., Ltd. \\
GranTokyo SOUTHTOWER, 1-9-2 Marunouchi, Chiyoda-ku, Tokyo 100-6640, Japan}
\author[label2]{Noriyoshi Sukegawa\corref{cor1}}
\address[label2]{Department of Information and System Engineering, \\
Faculty of Science and Engineering, Chuo University\\
1-13-27 Kasuga, Bunkyo-ku, Tokyo 112-8551, Japan}
\ead{sukegawa@ise.chuo-u.ac.jp}
\cortext[cor1]{Corresponding author}
\author[label3]{Yuichi Takano}
\address[label3]{School of Network and Information, Senshu University \\
2-1-1 Higashimita, Tama-ku, Kawasaki-shi, Kanagawa 214-8580, Japan}
\author[label4]{Jiro Iwanaga}
\address[label4]{Retty, Inc. \\
Takanawa Park Tower, 3-20-14 Higashi-Gotanda, Shinagawa-ku, Tokyo 141-0022, Japan}
\begin{abstract} 
This paper analyzes customer product-choice behavior based on the recency and frequency of each customer's page views on e-commerce sites. 
Recently, we devised an optimization model for estimating product-choice probabilities that satisfy monotonicity, convexity, and concavity constraints with respect to recency and frequency. 
This shape-restricted model delivered high predictive performance even when there were few training samples. However, typical e-commerce sites deal in many different varieties of products, so the predictive performance of the model can be further improved by integration of such product heterogeneity. 
For this purpose, we develop a novel latent-class shape-restricted model for estimating product-choice probabilities for each latent class of products. 
We also give a tailored expectation-maximization algorithm for parameter estimation. 
Computational results demonstrate that higher predictive performance is achieved with our latent-class model than with the previous shape-restricted model and common latent-class logistic regression. 
\end{abstract}


\end{frontmatter}



\section{Introduction}

Nowadays, a wide variety of products are browsed and purchased on e-commerce sites~\cite{TuKi15}. 
This enables the automated collection of clickstream data, which is a record of a visitor's page view (PV) history. 
Consequently, the analysis of clickstream data has been drawing intense research interest with respect to various topics, such as website browsing and navigation, Internet advertising, and online purchase behavior on e-commerce sites~\cite{BuSi09}. 
In particular, the present paper analyzes customer product-choice behavior based on clickstream data. 
The results of this research could be used to help visitors go to a target page on an e-commerce site and find the products they want. 
It could also be useful in demand forecasting for inventory management~\cite{HuMi14}.

The recency and frequency of a customer's previous purchases have been shown to be key indicators for forecasting repeat purchases~\cite{FaHa05a,FaHa05b,JeFa11,ReKu00,ReKu03}. 
In light of this fact, Iwanaga et al.~\cite{IwNi16} recently devised optimization models for estimating product-choice probabilities from the recency and frequency of each customer's previous PVs. 
These models exploit the properties of recency and frequency of PVs to enhance their predictive performance. 
In particular, the monotonicity--convexity--concavity (MCC) model estimates product-choice probabilities that satisfy monotonicity, convexity, and concavity constraints with respect to recency and frequency. 
Iwanaga et al.~\cite{IwNi16} demonstrated that higher predictive performance is achieved with MCC model than with the standard general-purpose methods for binary classification, namely, logistic regression and kernel-based support vector machines.

However, it is noteworthy that typical e-commerce sites deal in a wide variety of products. 
It is clear that purchase behavior toward products varies according to price range, purchase interval, and so on; nevertheless, such product heterogeneity is completely absent from the MCC model. 
For this reason, its predictive performance could be further improved by integrating product heterogeneity into the MCC model. 

For this purpose, we propose a latent-class shape-restricted model by applying latent-class modeling~\cite{HaMc02,LaHe68} to the MCC model. 
Our model classifies products into a specified number of latent classes and simultaneously estimates the product-choice probabilities for each latent class. 
To estimate the parameters of our model efficiently, we develop a tailored expectation-maximization (EM) algorithm~\cite{DeLa77,McKr07}. 
We assess the predictive performance of our model through computational experiments that use actual clickstream data. 

The contributions of the present paper are summarized as follows: 
\begin{itemize}
\item We establish a novel latent-class approach to estimating product-choice probabilities and we develop an EM algorithm to determine the associated parameter values. 
\item We verify through computational experiments that higher predictive performance is achieved with our model than with the MCC model~\cite{IwNi16} and latent-class logistic regression~\cite{Fo82,Ha79,KaRu89}. 
\item We closely examine customer product-choice behavior toward each latent class of products on the basis of the estimated product-choice probabilities. 
\end{itemize}

The remainder of this paper is organized as follows. 
Section~2 gives a brief review of related work. 
Section~3 presents the MCC model~\cite{IwNi16} for estimating product-choice probabilities from clickstream data. 
Section~4 develops our latent-class MCC model and EM algorithm. 
Section~5 evaluates the effectiveness of our model through computational experiments. 
Section~6 concludes with a brief summary of our work and a discussion of future research directions. 

\section{Related Work}
This section discusses some related works on the prediction of online purchasing behavior.
Then it gives a brief survey of the shape-restricted regression and latent-class regression. 

\subsection{Predicting online purchasing behavior}
One of the most active areas of clickstream research has been the analysis of online purchasing behavior of customers on e-commerce sites~\cite{BuSi09}. 
Moe and Fader~\cite{MoFa04} proposed a stochastic model for predicting online purchasing conversion rates based on an observed history of visits and purchases. 
Many other studies used logit and/or probit modeling based on various types of explanatory variables to predict online purchasing behavior~\cite{MoLi04,OlHo11,SaAs12,SiBu04,VaBu05}, whereas Boroujerdi et al.~\cite{BoMe14} applied several classification algorithms to predict customers' buying intentions.
However, these studies focused on predicting customer visits that culminated in purchases, and so they did not assign a purchase probability to each product. 

Although various studies have analyzed online product-choice behavior, most have emphasized more detailed data (e.g., multiplex data~\cite{ChFa12}, social media profiles~\cite{ZhPe13}, and product reviews and ratings~\cite{Qi14}) rather than clickstream data. 
In contrast to these studies, the present paper focuses on investigating the relationship between customer PVs and product-choice probabilities based on clickstream data. 
Our research will be of significant value to most e-commerce sites, for which the analysis of clickstream data is a major challenge. 

\subsection{Shape-restricted regression}
Shape-restricted regression, which has its origin in earlier work~\cite{Br55,Gr56,Hi54}, fits a nonparametric function to given data points under shape restrictions such as monotonicity, convexity, and concavity. 
Some popular examples include the estimation of utility/production/cost/profit functions in economics~\cite{GaGo84,Te96} and option pricing functions in finance~\cite{AiDu03}. 
Various algorithms have been developed for shape-restricted regression~\cite{Br58,Ch09,FrMa89,Ma13,RoWr88,WaGh12}, a special case of which is isotonic regression. 
This has many applications in statistics, operations research, and image processing~\cite{PaXu99}. 
Iwanaga et al.~\cite{IwNi16} recently used the maximum likelihood method to estimate product-choice probabilities subject to the monotonicity, convexity, and concavity constraints with respect to recency and frequency. 
Their shape-restricted model was a new effective application of shape-restricted regression to the analysis of clickstream data. 

\subsection{Latent-class regression}
Latent-class (or mixture) regression~\cite{WeDe94,WeDe02} is a traditional form of latent-class modeling~\cite{HaMc02,LaHe68}. 
It is aimed at classifying a sample into latent classes and simultaneously forming a regression model within these classes. 
There are two main algorithms for maximum likelihood estimation of latent-class regression models: the Newton--Raphson algorithm~\cite{Ev12} and the EM algorithm~\cite{DeLa77,McKr07}. 
While these models are frequently used in marketing and business research to represent consumer heterogeneity, we apply latent-class modeling to the classification of a wide variety of products on e-commerce sites. 
For this purpose, we develop a latent-class model for estimating the product-choice probabilities that satisfy monotonicity, convexity, and concavity constraints. 
To the best of our knowledge, none of the existing studies have considered a latent-class model under such a variety of shape restrictions. 
We also demonstrate that our latent-class model is superior to latent-class logistic regression~\cite{Fo82,Ha79,KaRu89} in terms of predictive performance based on clickstream data. 

\section{Monotonicity--convexity--concavity Model}
This section introduces the optimization models developed by Iwanaga et al.~\cite{IwNi16} for estimating product-choice probabilities from the recency and frequency of each customer's previous PVs. 

\subsection{Probability table}
Recency and frequency of PVs are defined for each customer--product pair. 
Roughly speaking, the \emph{recency} of customer $u$ with respect to product $v$ represents the time of the most recent visit of customer $u$ to the webpage of product $v$, while the \emph{frequency} represents the number of visits of customer $u$ to the webpage of product $v$. 

Let $I$ and $J$ denote finite sets of positive integers associated with the recency and frequency values, respectively. 
Then, the two-dimensional probability table is expressed as a matrix 
\[
X = (x_{ij})_{(i,j) \in I \times J} \in [0,1]^{I \times J},
\] 
where $x_{ij}$ indicates the probability that a product will be chosen (e.g., purchased) by a customer for recency $i$ and frequency $j$ of this customer--product pair. 
We will analyze customer product-choice behavior based on this probability table. 

It is relatively easy to calculate empirical product-choice probabilities by counting the number of purchases for each recency--frequency pair $(i,j) \in I \times J$ in past clickstream data.
In this case, however, if the element $x_{ij}$ is calculated from only a few training samples, it will not be particularly reliable. 
To resolve this problem, Iwanaga et al.~\cite{IwNi16} exploited the properties of recency and frequency of PVs in their optimization models . 

\subsection{Optimization models}
Let us specify the base date in past clickstream data; accordingly, we predict a customer's choice of products on the basis of information as of this date. 
For each $(i,j) \in I \times J$, we denote by $n_{ij}$ the number of customer--product pairs $(u,v)$ for which customer $u$ had recency $i$ and frequency $j$ for product $v$ as of the base date. 
We then set $q_{ij}$ to the number of those pairs that led to purchases since the base date. 
Given a probability table $X$, the occurrence probability of event $E := ((n_{ij},q_{ij});(i,j) \in I \times J)$ is expressed by the binomial distribution as follows:
\begin{align}\label{eq:prob1}
f(E;X) := \prod_{(i,j) \in I \times J} \binom{n_{ij}}{q_{ij}} (x_{ij})^{q_{ij}} (1 - x_{ij})^{n_{ij} - q_{ij}}. 
\end{align}
The following optimization models maximize the log-likelihood function $\log(f(E;X))$ after omitting its constant terms. 

It is assumed that the product-choice probability increases with recency and frequency because they reflect growing customer interest in a particular product. 
Accordingly, the monotonicity model~\cite{IwNi16} estimates a probability table $X$ under the following monotonicity constraints: 
\begin{equation}
\begin{array}{|cll}
\ds \mathop{\mbox{maximize}} 
& \ds \sum_{(i,j) \in I \times J} \left( q_{ij} \log x_{ij} + (n_{ij} - q_{ij}) \log (1 - x_{ij}) \right) \cr
\mbox{subject~to} 
& \ds x_{ij} \le x_{i+1,j}~~~~((i,j) \in I \times J,~i \le |I|-1), \cr
& \ds x_{ij} \le x_{i,j+1}~~~~((i,j) \in I \times J,~j \le |J|-1), \cr 
& \ds 0 < x_{ij} < 1~~~~((i,j) \in I \times J). \cr 
\end{array}
\label{form:mono}
\end{equation}

Additionally, as the customer's most recent visit becomes longer ago, the variation in recency value should have less effect. 
In other words, the larger the recency value, the larger the increment in the purchase probability. 
Moreover, as the number of visits of a particular customer grows, the effect of each visit will be smaller; that is, the larger the frequency value, the smaller the increment in the purchase probability. 
The MCC model~\cite{IwNi16} includes these properties via the convexity--concavity constraints on product-choice probability: 
\begin{equation}
\begin{array}{|cll}
\ds \mathop{\mbox{maximize}} 
& \ds \sum_{(i,j) \in I \times J} \left( q_{ij} \log x_{ij} + (n_{ij} - q_{ij}) \log (1 - x_{ij}) \right) \cr
\mbox{subject~to} 
& \ds x_{ij} \le x_{i+1,j} ~~~~ ((i,j) \in I \times J,~i \le |I|-1), \cr
& \ds x_{ij} \le x_{i,j+1} ~~~~ ((i,j) \in I \times J,~j \le |J|-1), \cr
& \ds x_{i+1,j} - x_{i,j} \le x_{i+2, j} - x_{i+1, j} ~~~~ ((i,j) \in I \times J,~i \le |I|-2), \cr
& \ds x_{i,j+1} - x_{ij} \ge x_{i,j+2} - x_{i,j+1} ~~~~ ((i,j) \in I \times J,~j \le |J|-2), \cr 
& \ds 0 < x_{ij} < 1 ~~~~ ((i,j) \in I \times J). \cr 
\end{array}
\label{form:mcc}
\end{equation}

It should be noted here that these optimization models apply the same probability table to all products; however, distinct categories of products can elicit different purchasing behavior. 
Hence, in the next section, we consider developing multiple probability tables that reflect the diversity of products. 

\section{Latent-class Model}\label{sec:lcm}
This section presents our latent-class model for estimating product-choice probabilities. 
It then describes an EM algorithm for setting the parameters of our model. 

\subsection{Latent-class modeling of probability table}\label{subsec:lcmpt}
Suppose that we are given a set $K$ of product categories. 
For each $(i,j,k) \in I \times J \times K$, we define $n_{ijk}$ as the number of customer--product pairs that had recency $i$ and frequency $j$ for a product of category $k$ as of the base date. 
We also set $q_{ijk}$ to the number of those pairs that were accompanied by purchases since the base date. 

The simplest way to distinguish between product categories would be to create a different probability table for each category. 
However, such an approach would lead to an unreliable probability table if some categories have only a small number of training samples. 
For this reason, we aggregate homogeneous categories of products in the same manner as latent-class regression~\cite{WeDe94,WeDe02}. 

Specifically, we introduce a set $S$ of latent classes such that $|S| < |K|$. 
The size of class $s \in S$ is $\pi_s$ and satisfies 
\begin{align} \label{con:size}
\sum_{s \in S} \pi_s = 1~~~~\mbox{and}~~~~\pi_s > 0~~~~(s \in S). 
\end{align}
A two-dimensional probability table is then defined for each class $s \in S$: 
\[
X_s = (x_{ijs})_{(i,j) \in I \times J} \in [0,1]^{I \times J},
\] 
where $x_{ijs}$ indicates the probability that a product in class $s$ will be chosen by a customer when the corresponding recency and frequency values are $i$ and $j$, respectively. 

As with Eq.~\eqref{eq:prob1}, the conditional probability of event $E_k := ((n_{ijk},q_{ijk});(i,j) \in I \times J)$, given that category $k$ belongs to class $s$, is written as 
\[
f(E_k;X_s) = \prod_{(i,j) \in I \times J} \binom{n_{ijk}}{q_{ijk}} (x_{ijs})^{q_{ijk}} (1 - x_{ijs})^{n_{ijk} - q_{ijk}}.
\]
By mixing these probabilities with weight $\pi_s$, the unconditional probability of event $E_k$ becomes 
\[
\sum_{s \in S} \pi_s f(E_k;X_s). 
\]
The posterior probability of category $k$ belonging to class $s$ is given by Bayes' rule as follows: 
\begin{align} \label{eq:postprob}
\frac{\pi_s f(E_k;X_s)}{\sum_{s \in S} \pi_s f(E_k;X_s)}. 
\end{align}

\subsection{EM algorithm}\label{subsec:ema}
To determine the parameters of our latent-class model, we develop a tailored EM algorithm. 
We begin by introducing unknown membership variables $z_{ks} \in [0,1]$ for $(k,s) \in K \times S$; that is, $z_{ks} = 1$ if category $k$ belongs to class $s$; otherwise, $z_{ks} = 0$. 
Then, the complete-data log-likelihood function is expressed as 
\begin{align}
  & \log \left( \prod_{(k,s) \in K \times S} (\pi_s f(E_k;X_s))^{z_{ks}} \right) \notag \\
= & \sum_{(k,s) \in K \times S} z_{ks} \log f(E_k;X_s) + \sum_{(k,s) \in K \times S} z_{ks} \log \pi_s. \label{eq:complog}
\end{align}

The EM algorithm starts with some initial estimate of membership variables $\hat{z}_{ks}$ for $(k,s) \in K \times S$. 
It then repeats the E-step (expectation step) and M-step (maximization step) to maximize the log-likelihood function~\eqref{eq:complog}. 

The M-step substitutes $\hat{z}_{ks}$ into the log-likelihood function~\eqref{eq:complog} and then maximizes it with respect to other decision variables. 
Specifically, the latent-class size is determined by maximizing $\sum_{(k,s) \in K \times S} \hat{z}_{ks} \log \pi_s$ subject to constraints~\eqref{con:size}. 
The method of Lagrange multipliers yields 
\begin{align}\label{eq:Estep1}
\hat{\pi}_s \leftarrow \frac{\sum_{k \in K} \hat{z}_{ks}}{|K|}
\end{align}
for each $s \in S$.

Next, the product-choice probabilities are determined by maximizing $\sum_{(k,s) \in K \times S} \hat{z}_{ks} \log f(E_k;X_s)$ subject to the monotonicity and convexity--concavity constraints. 
This optimization problem can be decomposed into one for each $s \in S$; consequently, we solve 
\begin{equation}
\begin{array}{|cll}
\ds \mathop{\mbox{maximize}} 
& \ds \sum_{(i,j,k) \in I \times J \times K} \hat{z}_{ks} \Bigl( q_{ijk} \log x_{ijs}+ (n_{ijk} - q_{ijk}) \log (1 - x_{ijs}) \Bigr) \cr
\mbox{subject~to} 
& \ds x_{ijs} \le x_{i+1,j,s}~~~~((i,j) \in I \times J,~i \le |I|-1), \cr
& \ds x_{ijs} \le x_{i,j+1,s}~~~~((i,j) \in I \times J,~j \le |J|-1), \cr 
& \ds x_{i+1,j,s} - x_{ijs} \le x_{i+2,j,s} - x_{i+1,j,s} ~~~~ ((i,j) \in I \times J,~i \le |I|-2), \cr
& \ds x_{i,j+1,s} - x_{ijs} \ge x_{i,j+2,s} - x_{i,j+1,s} ~~~~ ((i,j) \in I \times J,~j \le |J|-2), \cr 
& \ds 0 < x_{ijs} < 1~~~~((i,j) \in I \times J) \cr 
\end{array}
\label{form:mcc2}
\end{equation}
to find a solution $\hat{X}_s = (\hat{x}_{ijs})_{(i,j) \in I \times J}$ for each $s \in S$. 
This optimization problem is similar to problem~\eqref{form:mcc} and maximizes a concave function subject to linear constraints; hence, it can be solved exactly and efficiently with standard nonlinear optimization software.

After that, the E-step calculates the expected value of each membership variable based on the current estimates of the other variables. 
This amounts to assigning the posterior probability~\eqref{eq:postprob} to the membership variable 
\begin{align}\label{eq:Mstep}
\hat{z}_{ks} \leftarrow \frac{\hat{\pi}_s f(E_k;\hat{X}_s)}{\sum_{s \in S} \hat{\pi}_s f(E_k;\hat{X}_s)}
\end{align}
for each $(k,s) \in K \times S$. 
The E-step and M-step are repeated until a termination condition is satisfied. 

Our EM algorithm for estimating the latent-class MCC model is summarized as follows:
\begin{description}
\item[Step 0]{\sffamily (Initialization)}~
Set $\hat{z}_{ks}$ for $(k,s) \in K \times S$ as initial estimates, and go to Step 2. 
\item[Step 1]{\sffamily (E-Step)}~
Update $\hat{z}_{ks}$ according to Eq.~\eqref{eq:Mstep} for $(k,s) \in K \times S$. 
\item[Step 2]{\sffamily (M-Step)}~
Update $\hat{\pi}_s$ according to Eq.~\eqref{eq:Estep1} for $s \in S$. 
Update $\hat{X}_s$ with a solution to problem~\eqref{form:mcc2} for $s \in S$. 
\item[Step 3]{\sffamily (Termination Condition)}~
Terminate the algorithm if a termination condition is satisfied. 
Otherwise, return to Step 1.
\end{description}

\section{Computational Experiments}\label{sec:ce}

The computational results reported in this section evaluate the effectiveness of our latent-class model for estimating product-choice probabilities. 

\subsection{Clickstream data}\label{sec:cd}

We used actual clickstream data provided by Recruit Lifestyle Co., Ltd.,\footnote{\url{http://www.recruit-lifestyle.co.jp/english/}} a leading Japanese company in products and services for food, beauty, travel, mail order, group discount tickets, and other areas of daily consumption. 
As will be seen in Section~\ref{sec:aelc}, we considered 56 categories of products that were purchased relatively frequently on the targeted e-commerce site. 
The clickstream data were collected during August--October 2015 from roughly 4 million customers, the top 1\% of which in terms of the number of purchases were excluded as outliers. 
Each transaction in the data corresponds to either a ``page view (PV)" or a ``purchase'' and contains information such as time, customer ID, and product ID.

We prepared the following six features representing recency and frequency based on PV, session, and day:
\begin{description}
\item[ViewR] Recency that is measured in PVs ($|I|=24$) 
\item[SesR] Recency that is measured in sessions ($|I|=12$) 
\item[DayR] Recency that is measured in days ($|I|=24$) 
\item[ViewF] Frequency that is measured in PVs ($|J|=16$) 
\item[SesF] Frequency that is measured in sessions ($|J|=8$) 
\item[DayF] Frequency that is measured in days ($|J|=8$)
\end{description}
For instance, if the most recent PV for a customer--product pair was made $m~(\ge 1)$ days ago, the DayR value was set to $i = \max\{25 - m, 1\}$. 
If $n~(\ge 1)$ PVs were made for a customer--product pair, the ViewF value was set to $j = \min\{n, 16\}$. 
These recency and frequency values were calculated from the clickstream data observed in the four weeks before the base date. 
Also, the threshold values (i.e., $|I|$ and $|J|$) of recency and frequency were determined such that the recency/frequency values were rounded up/down for less than 5\% of all customer--product pairs. 

\subsection{Experimental design}\label{sec:ed}

We compare the predictive performances of the following models through the top-$N$ purchase prediction: 
\begin{description}
\item[MCC($1$)] MCC model~\eqref{form:mcc}, which develops only one probability table and applies it to all products. 
\item[MCC($56$)] MCC model~\eqref{form:mcc}, which develops 56 probability tables separately for the various product categories. 
\item[LCMCC($|S|$)] Our latent-class MCC model (see Section~\ref{sec:lcm}); $|S|$ is the number of latent classes.
\item[LCLR($|S|$)] Latent-class logistic regression that predicts a product purchase from two input variables, namely the recency and frequency of each customer--product pair; $|S|$ is the number of latent classes.
\end{description}
The model parameters of LCMCC($|S|$) were tuned by our EM algorithm (see Section~\ref{subsec:ema}), and those of LCLR($|S|$) were tuned by the standard EM algorithm~\cite{WeDe94,WeDe02}. 
The two EM algorithms were terminated if the increment in the log-likelihood from the previous iteration was sufficiently small, where the maximum number of iterations was set to 10. 
We repeated this EM-algorithm process 10 times with random initial estimates and chose the best obtained estimates in log-likelihood. 
The associated optimization problems (e.g., \eqref{form:mcc} and \eqref{form:mcc2}) were solved using Numerical Optimizer V17, a numerical optimization software developed by NTT DATA Mathematical Systems Inc.\footnote{\url{http://www.msi.co.jp/english/}} 
Here, the inequality constraint $0 < x_{ij} <1$ was replaced with $\varepsilon \le x_{ij} \le 1-\varepsilon$ with $\varepsilon=10^{-5}$. 
All the computations were performed on a Windows computer with an AMD Opteron 4133 CPU (2.80~GHz) and 128~GB of memory. 

In our top-$N$ purchase prediction, the purchase data collected in September 2015 were used in the training phase and those collected in October 2015 were used in the test phase. 
The training phase estimates two-dimensional probability tables. 
The base date was first set to September 3, and the recency and frequency values and the purchase data of the base date were gathered for every customer--product pair. 
Similar data were collected by moving the base date one day at a time from September 3 to 30. 
This data set was used as a training set that consists of 43,566 purchase samples and 96,495,418 non-purchase samples, in which each customer--product pair corresponds to one sample. 
To examine the relationship between the number of training samples and the predictive performance, two additional training sets were generated by randomly selecting samples from the original training set at sampling rates of $1\%$ or $10\%$. 
These synthetic training sets are referred to as the ``1\%-sampled'' and ``10\%-sampled'' sets, and the original one is referred to as the ``100\%-sampled'' set as required. 
Two-dimensional probability tables were estimated from these training sets. 

The test phase evaluates the predictive performances of the two-dimensional probability tables. 
The base date was first set to October 1, and the product-choice probabilities were computed by entering the recency and frequency values to the probability tables. 
Then $N$ products were selected for each customer in descending order of the computed product-choice probabilities. 
Here, ViewF was used for tie-breaks when some products had the same product-choice probability. 
For customers who viewed fewer than $N$ products in the previous four weeks, those products were all selected. 
This process was repeated by moving the base date from October 1 to 28. Consequently, 28 test sets were completed, each of which contained around 1,214 purchase samples and 3,316,018 non-purchase samples on average. 

In each test set, we calculated the following criteria for every customer: 
\begin{itemize}
\item[] $\ds \text{Recall} = \frac{\text{\#(selected and purchased products)}}{\text{\#(purchased products)}}$,
\item[] $\ds \text{Precision} = \frac{\text{\#(selected and purchased products)}}{\text{\#(selected products)}}$,
\item[] $\ds \text{F1 score} = \frac{2\cdot\text{Recall}\cdot\text{Precision}}{\text{Recall}+\text{Precision}}$,
\end{itemize}
where $\#(\,\cdot\,)$ stands for the number of corresponding products. 
Here, the purchased products were restricted to those viewed by the targeted customer in the previous four weeks. 
These F1 scores were averaged over customers, and they were further averaged over the 28 test sets as needed. 

\subsection{Combinations of recency and frequency features}\label{sec:crff}

\begin{figure}[tb]
\begin{minipage}{0.48\textwidth}
\centering
\begin{tabular}{ccc}
\includegraphics[clip,width=200pt]{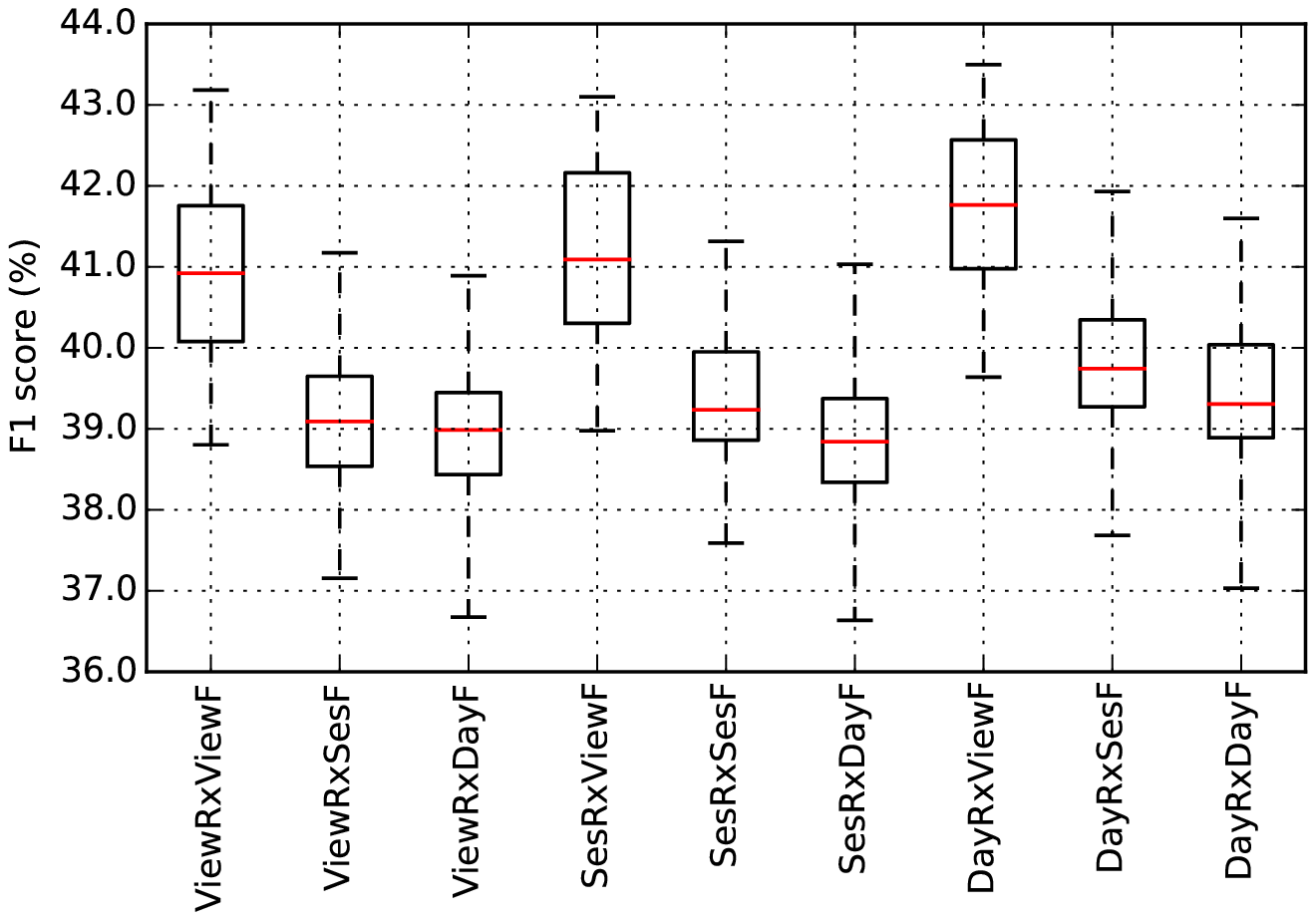}\\
{\small (a) $N=3$}\\
\includegraphics[clip,width=200pt]{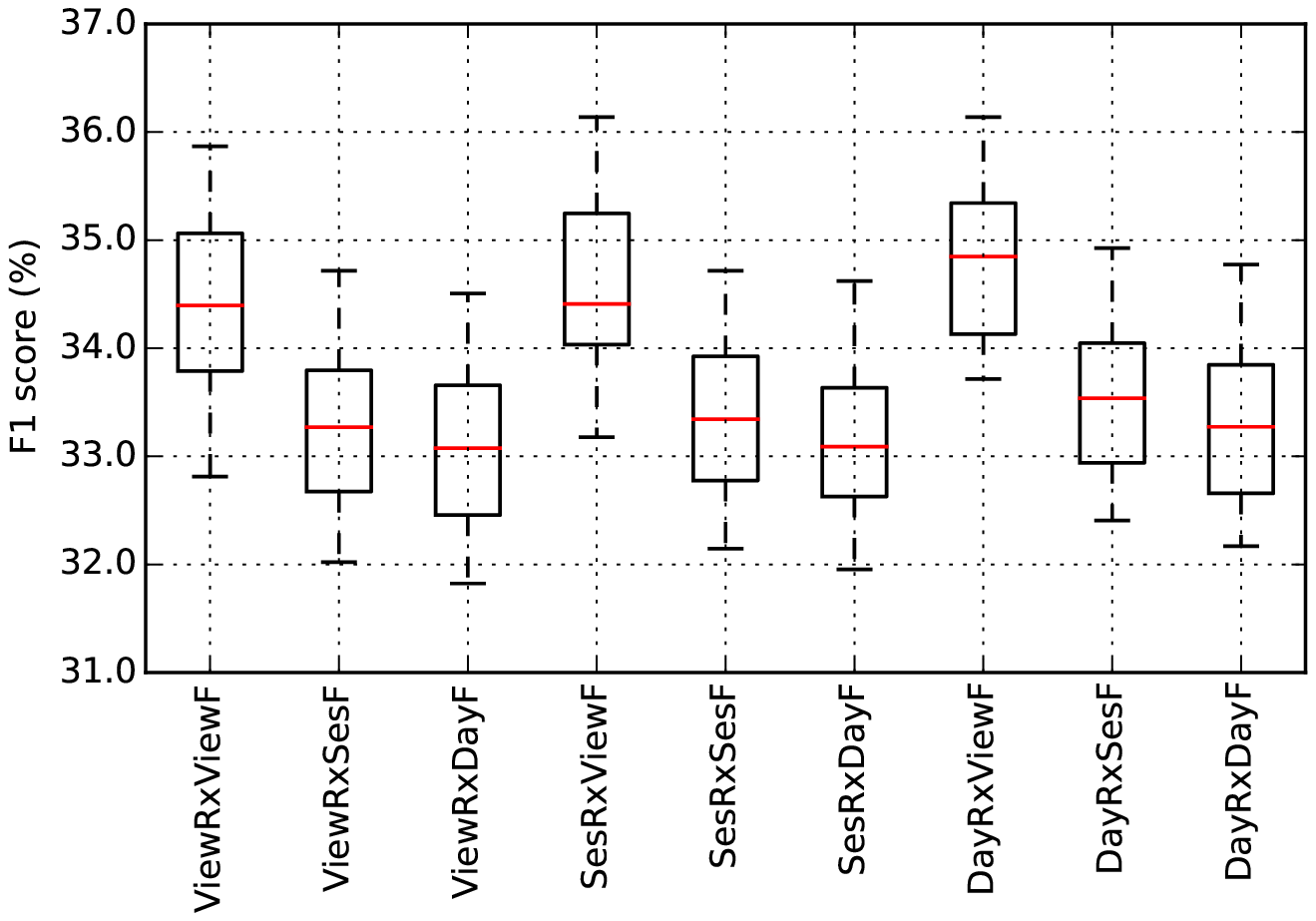}\\
{\small (b) $N=5$}\\
\includegraphics[clip,width=200pt]{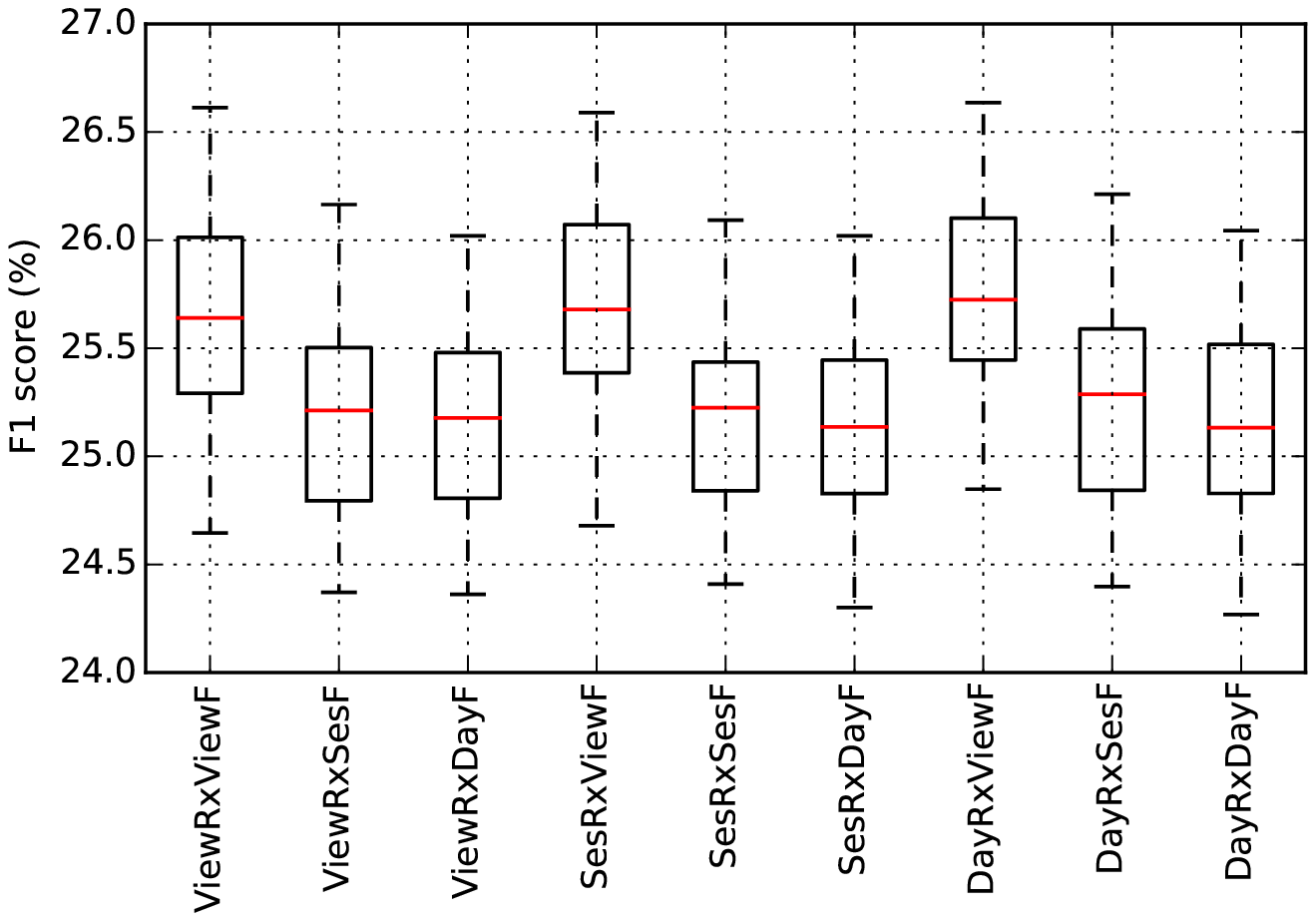}\\
{\small (c) $N=10$}
\end{tabular}
\caption{F1 scores of MCC(1) based on nine combinations of recency and frequency features
 \label{fig:1}}
\end{minipage}\hfill
\begin{minipage}{0.48\textwidth}
\centering
\begin{tabular}{ccc}
\includegraphics[clip,width=200pt]{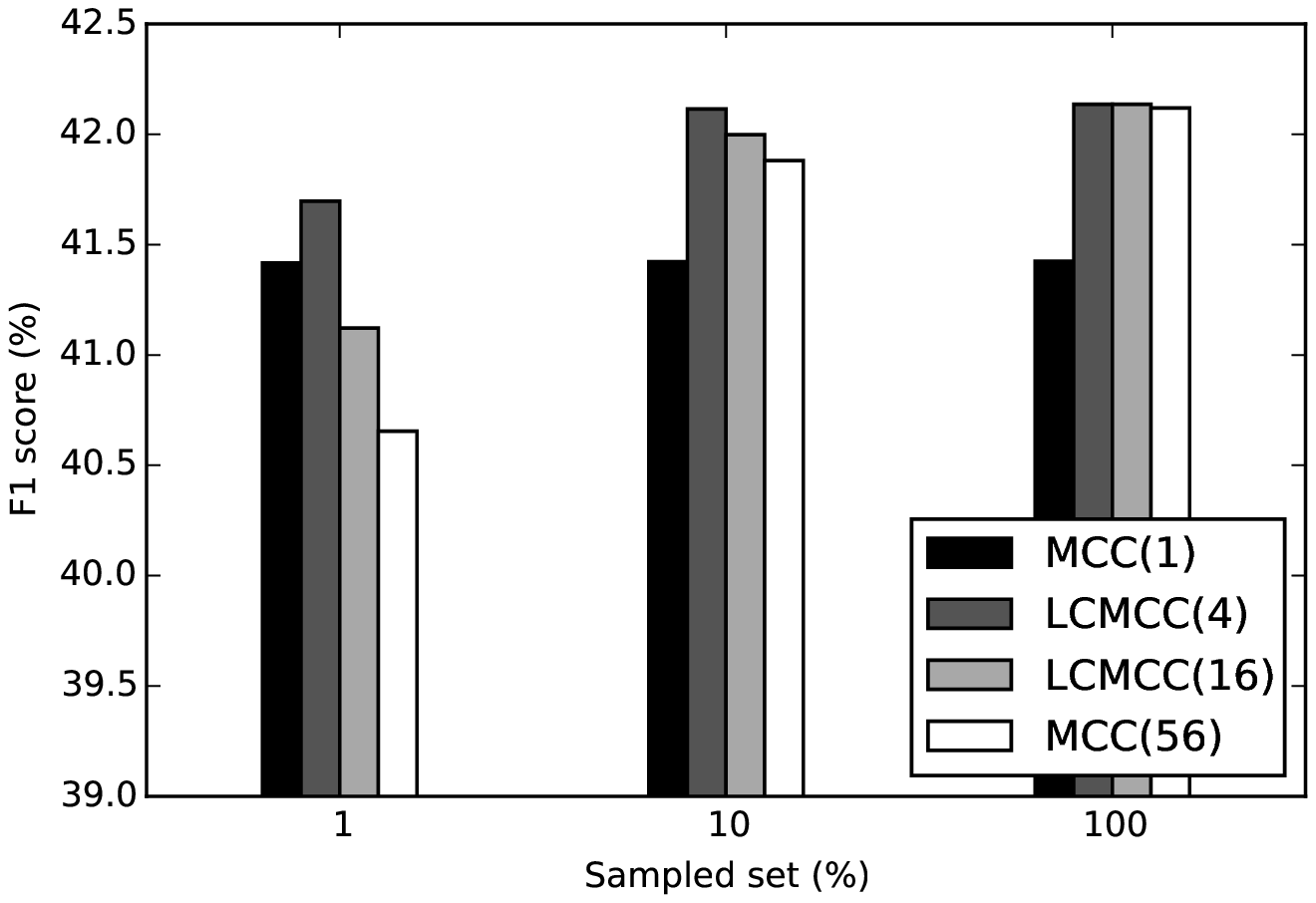}\\
{\small (a) $N=3$}\\
\includegraphics[clip,width=200pt]{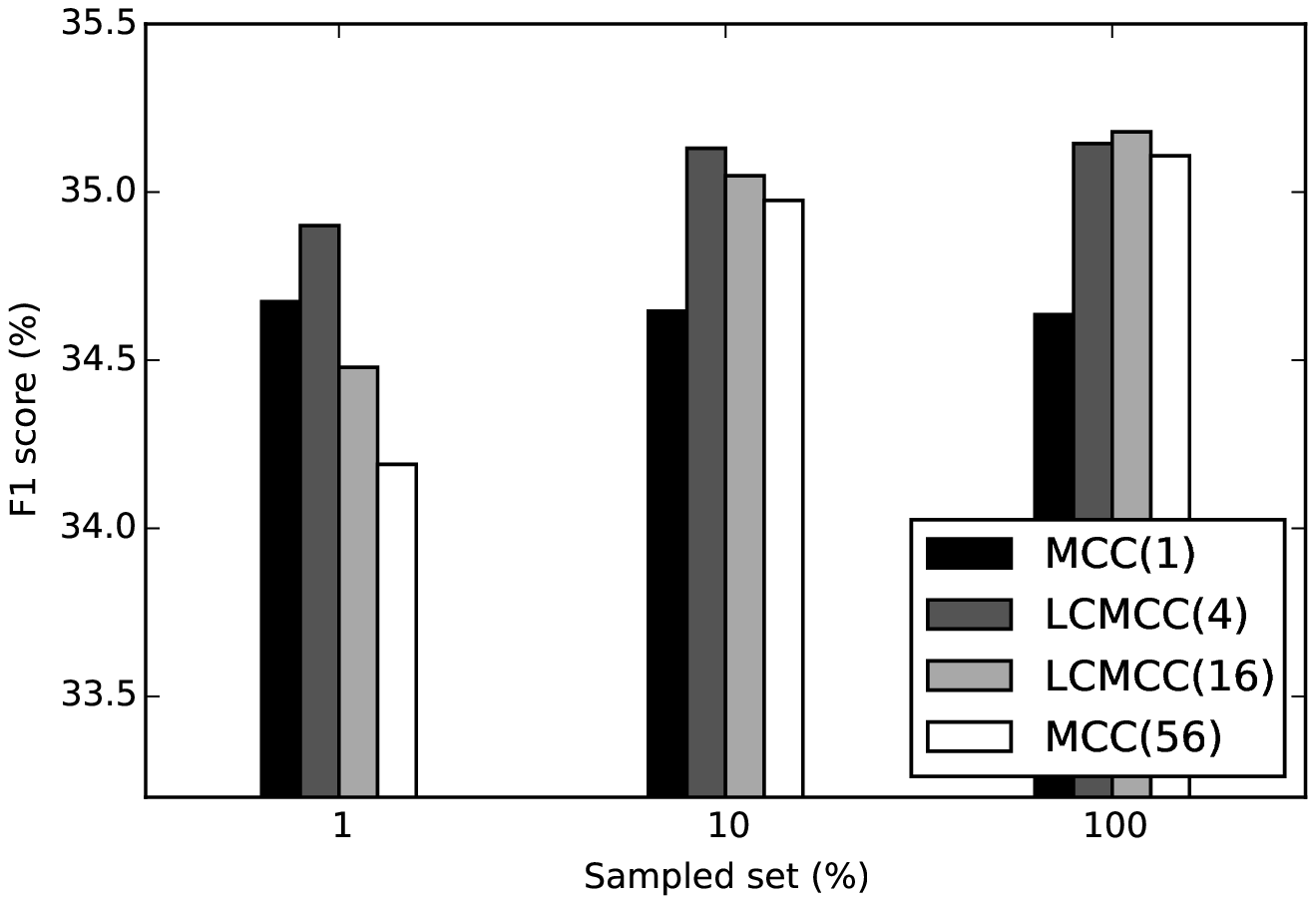}\\
{\small (b) $N=5$}\\
\includegraphics[clip,width=200pt]{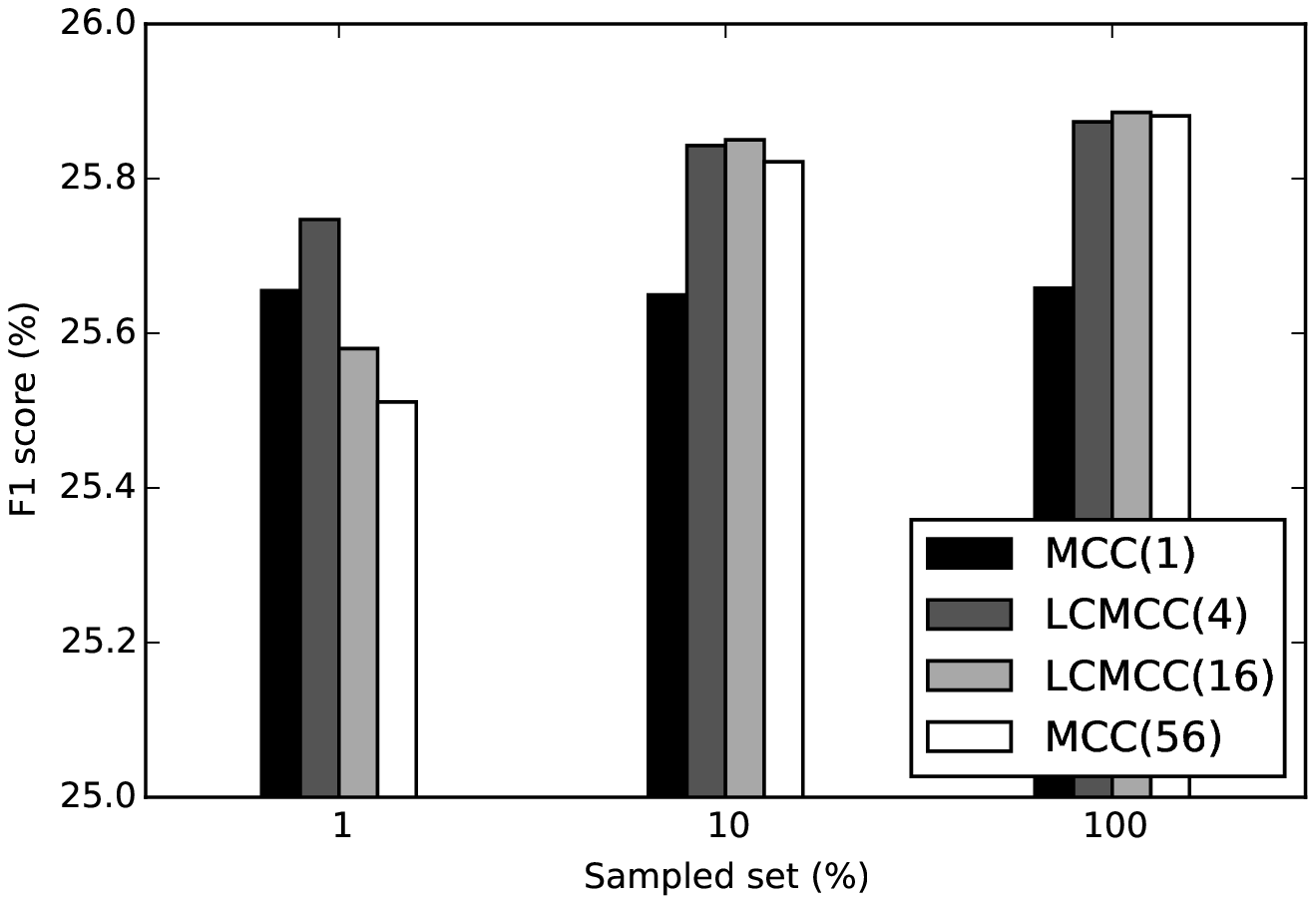} \\
{\small (c) $N=10$}
\end{tabular}
\caption{F1 scores of MCC and LCMCC models for sampled training sets \label{fig:2}}
\end{minipage}
\end{figure}

We begin by finding the best combination of recency and frequency features for estimating the product-choice probability. 
The box plots in Figure~\ref{fig:1} display the distribution of the F1 scores of MCC(1) for the 28 test sets, where the number of selected products is $N \in \{3,5,10\}$. 
Here, the nine combinations of recency features \{ViewR, SesR, DayR\} and frequency features \{ViewF, SesF, DayF\} were tested. 
It is clear from Figure~\ref{fig:1} that the use of ViewF improved the F1 score significantly. 
We can also see that DayR$\times$ViewF was the best combination for all $N \in \{3,5,10\}$. 
In view of these results, we employ this combination of DayR and ViewF in the following computational experiments. 

\subsection{Effectiveness of latent-class model}\label{sec:elcmpt}

We examine the effectiveness of our latent-class modeling for the MCC model. 
Figure~\ref{fig:2} shows the F1 scores of MCC(1), LCMCC(4), LCMCC(16), and MCC(56) for the 1\%-, 10\%-, and 100\%-sampled sets, where the number of selected products is $N \in \{3,5,10\}$. 

MCC(1) is less likely to overfit a small number of training samples because it uses only one probability table. 
For this reason, MCC(1) had roughly the same F1 score for all training sets. 
By contrast, MCC(56) creates 56 probability tables in accordance with the number of product categories. 
As a result, the predictive performance of MCC(56) was appreciably worse for the 1\%-sampled set but improved as the number of training samples increased. 

Meanwhile, our four-latent-class model, LCMCC(4), clearly achieved the highest predictive performance of all the models for the 1\%-sampled set. 
LCMCC(4) still outperformed MCC(1) and MCC(56) for the 10\%-sampled set. 
In the case of the 100\%-sampled set, LCMCC(4) substantially outperformed MCC(1) and was also competitive with MCC(56). 
Our 16-latent-class model, LCMCC(16), performed poorly for the 1\%-sampled set but delivered the best predictive performance of all the models for the 100\%-sampled set. 
These results verify that our latent-class model successfully enhanced the predictive performance of the previous MCC model by clustering 56 product categories into a smaller number of latent classes. 

\subsection{Comparison with latent-class logistic regression}\label{sec:clclr}

\begin{figure}[tb]
\centering
\begin{tabular}{ccc}
\includegraphics[clip,width=150pt]{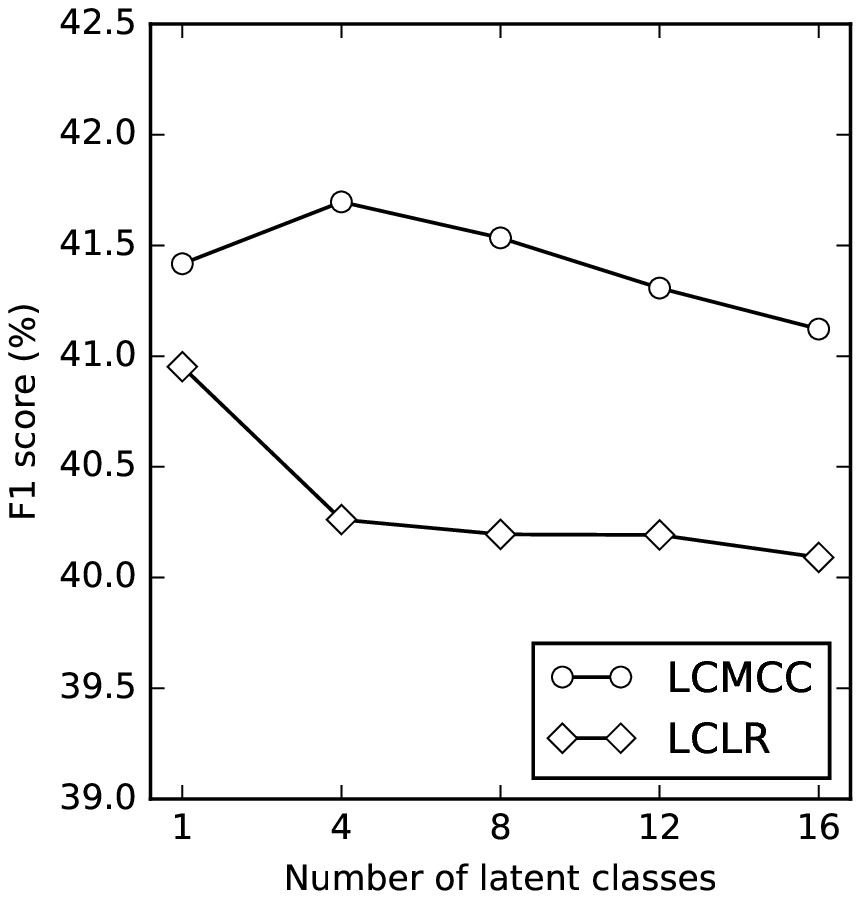}	&\includegraphics[clip,width=150pt]{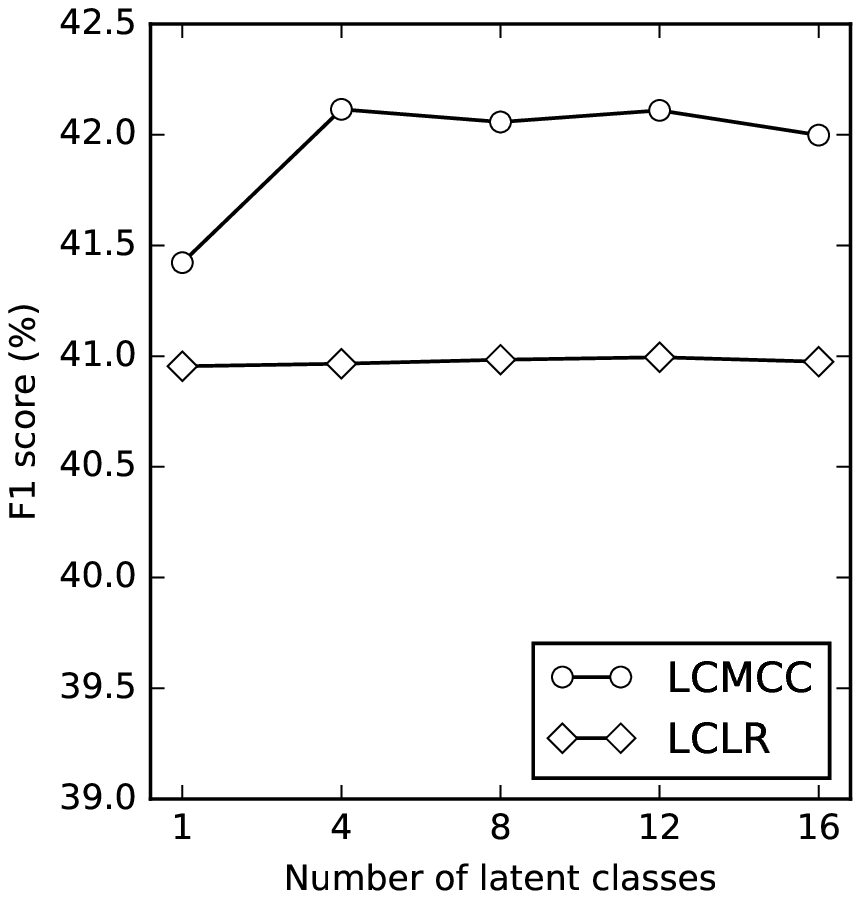}	&\includegraphics[clip,width=150pt]{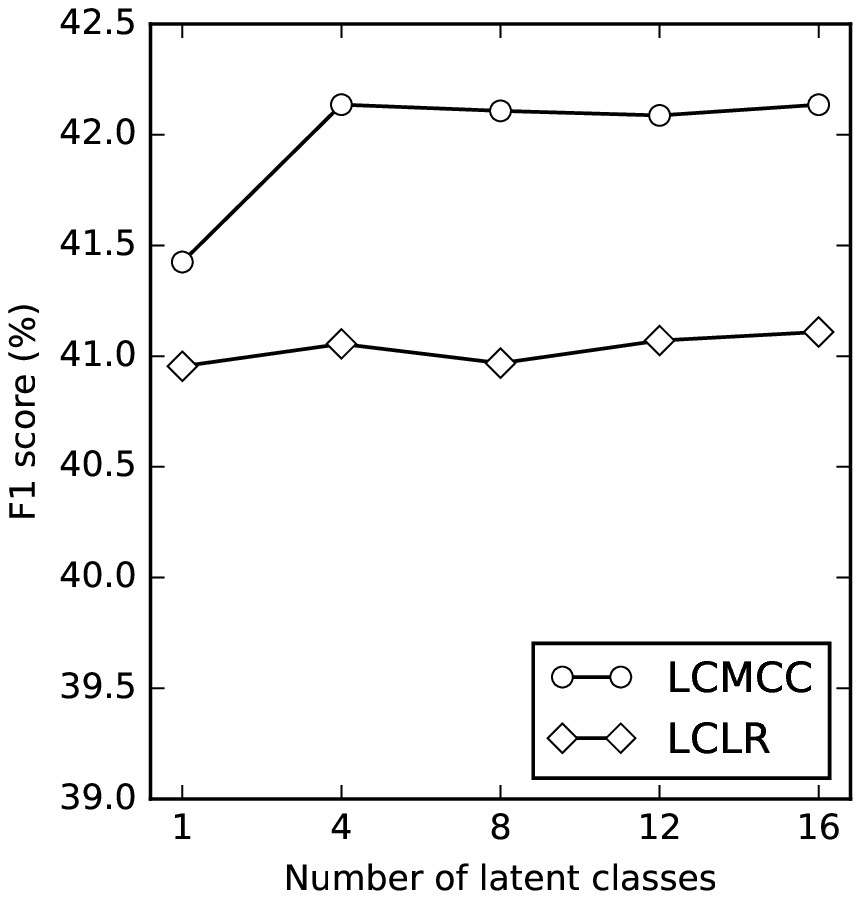}	\\
{\small (a) $N = 3$, $1\%$-sampled}&{\small (b) $N = 3$, $10\%$-sampled}&{\small (c) $N = 3$, $100\%$-sampled}\\
\includegraphics[clip,width=150pt]{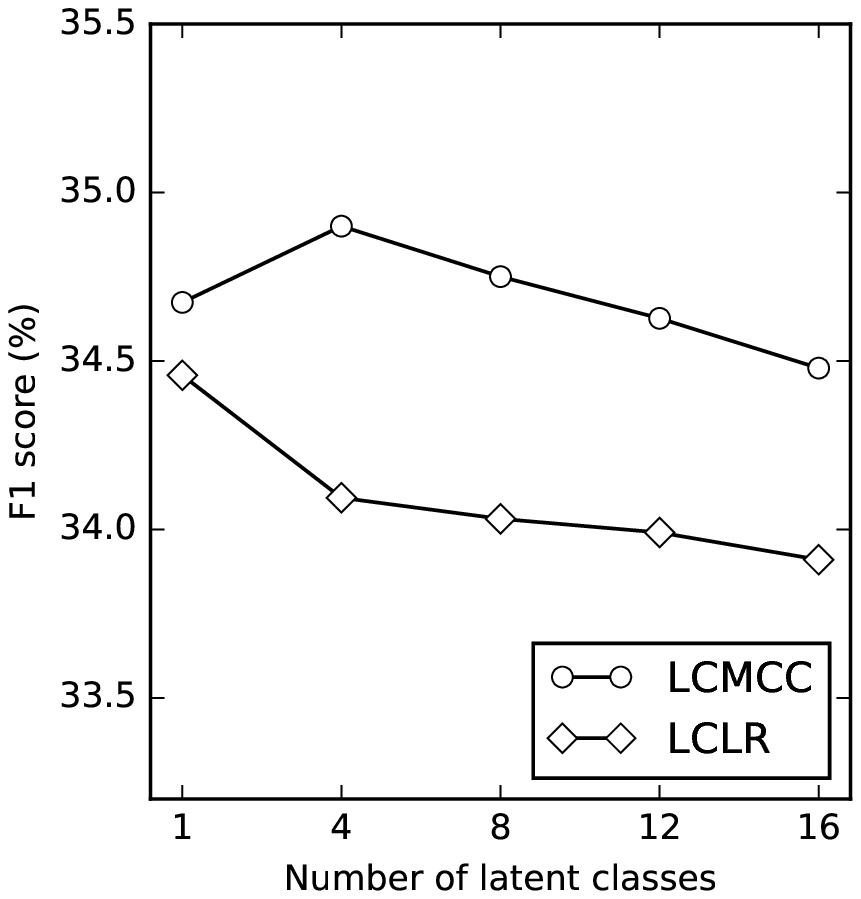}	&\includegraphics[clip,width=150pt]{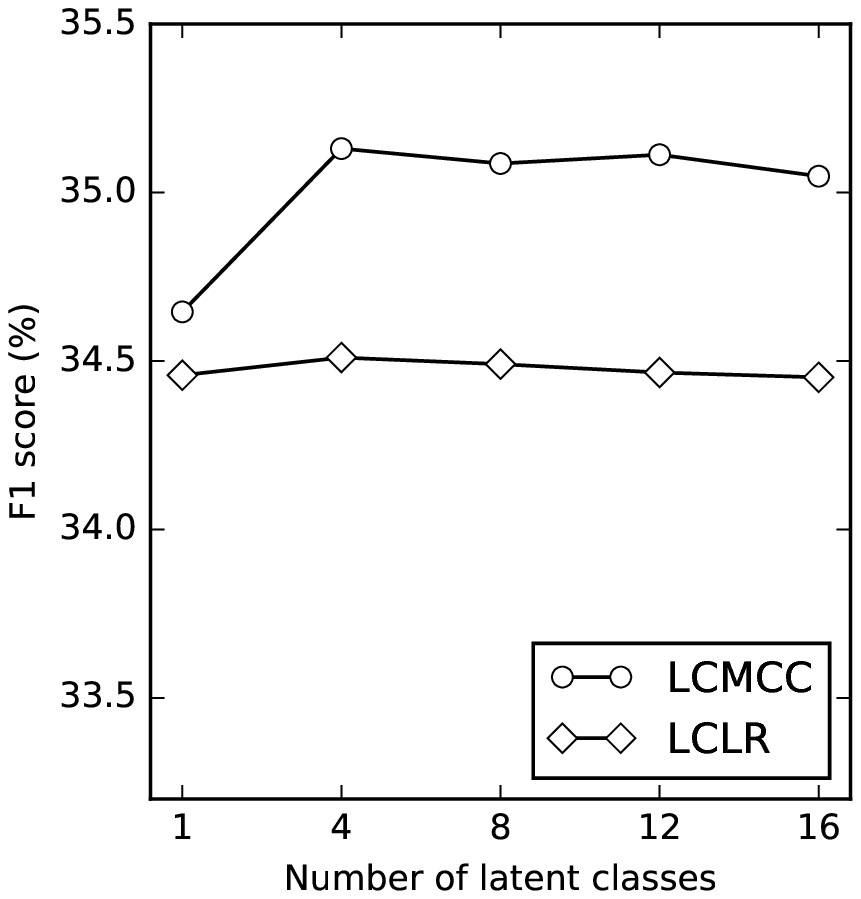}	&\includegraphics[clip,width=150pt]{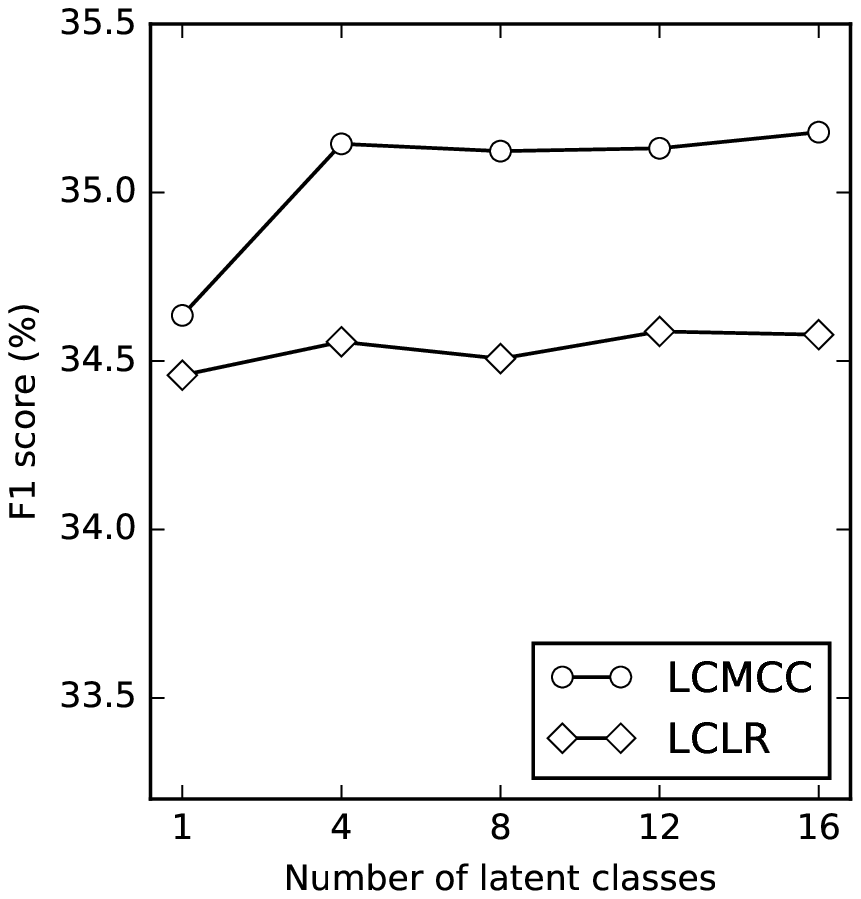}	\\
{\small (d) $N = 5$, $1\%$-sampled}&{\small (e) $N = 5$, $10\%$-sampled}&{\small (f) $N = 5$, $100\%$-sampled}\\
\includegraphics[clip,width=150pt]{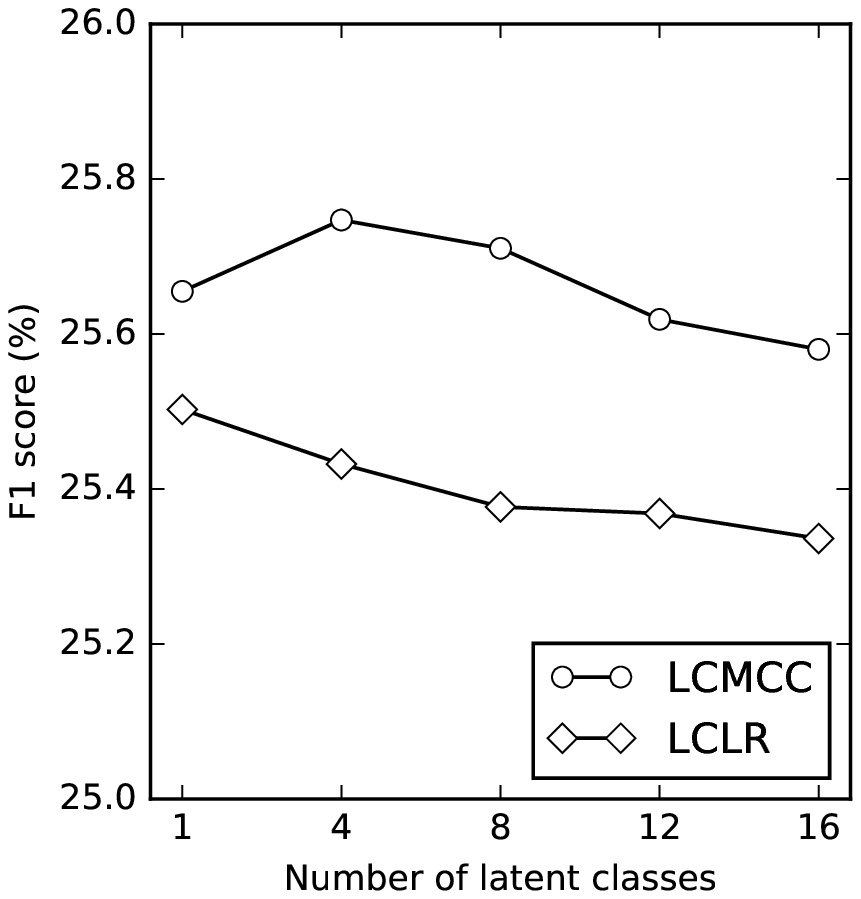}	&\includegraphics[clip,width=150pt]{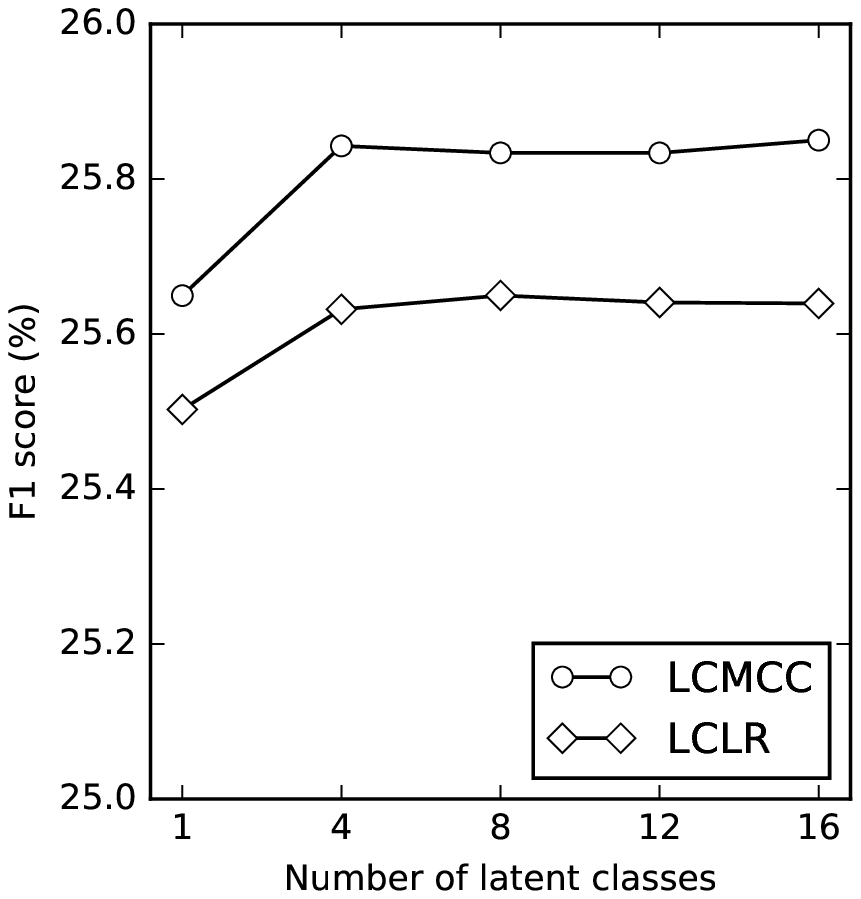}	&\includegraphics[clip,width=150pt]{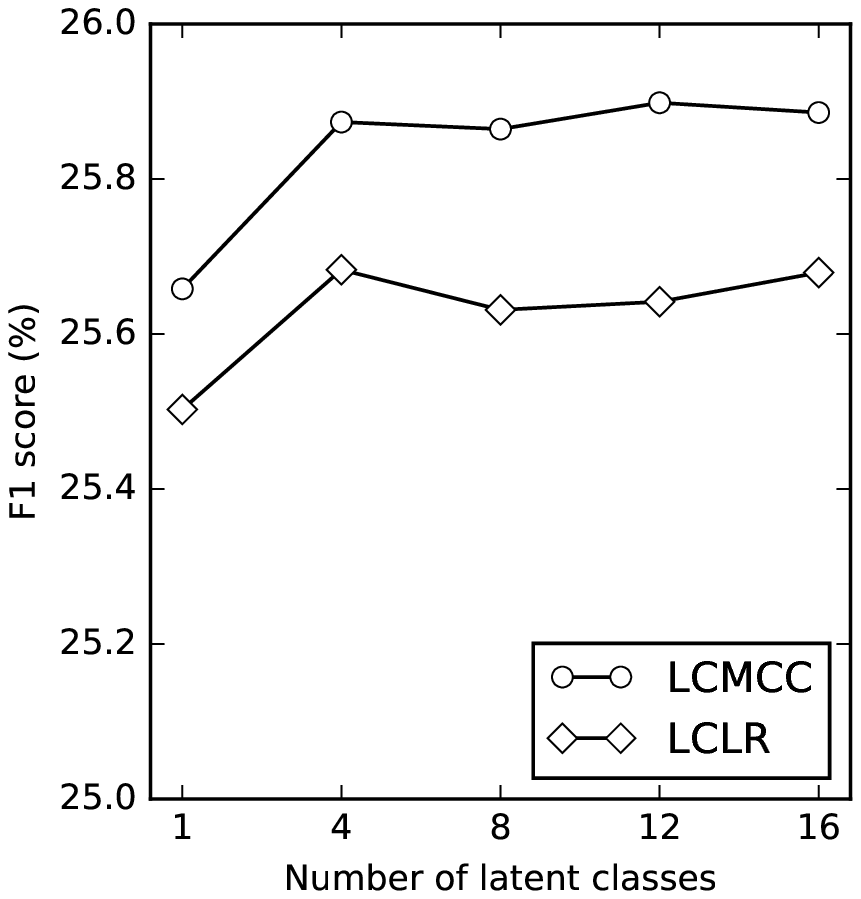}	\\ 
{\small (g) $N = 10$, $1\%$-sampled}&{\small (h) $N = 10$, $10\%$-sampled}&{\small (i) $N = 10$, $100\%$-sampled}\\
\end{tabular}
\caption{F1 scores of LCMCC and LCLR models for sampled training sets \label{fig:3}}
\end{figure}

We compare the predictive performance of our latent-class model with that of latent-class logistic regression. 
Figure~\ref{fig:3} shows the F1 scores of LCMCC($|S|$) and LCLR($|S|$) for the 1\%-, 10\%-, and 100\%-sampled sets, where the number of latent classes is $|S| \in \{1,4,8,12,16\}$ and the number of selected products is $N \in \{3,5,10\}$. 

Figure~\ref{fig:3} shows that the predictive performance of LCMCC($|S|$) was significantly higher than that of LCLR($|S|$) for all $|S| \in \{1,4,8,12,16\}$. 
In what follows, we closely examine the results for each sampled training set. 

We first focus on the results for the 1\%-sampled set. 
In this case, the predictive performance of LCLR($|S|$) decreased with the number $|S|$ of latent classes because of a shortage of training samples. 
By contrast, the F1 scores of LCMCC(4) and LCMCC(8) were higher than that of LCMCC(1). 
In other words, our LCMCC model has the potential to improve the predictive performance for even small-sample training sets. 

In the case of the 10\%-sampled set, the F1 score of LCMCC($4$) was much higher than that of LCMCC($1$), whereas the difference in the F1 score between LCLR($1$) and LCLR($4$) was relatively small. 
We can also see that the overall predictive performance was improved by employing the 10\%-sampled set instead of the 1\%-sampled set. 
Although similar results were obtained in the case of the 100\%-sampled set, it is noteworthy that the predictive performance of LCMCC($16$) for the 100\%-sampled set was slightly better than that for the 10\%-sampled set (see, e.g., Figure~\ref{fig:3}(e) and (f)). 

\begin{figure}[tb]
\centering
\begin{tabular}{ccc}
\includegraphics[clip,width=150pt]{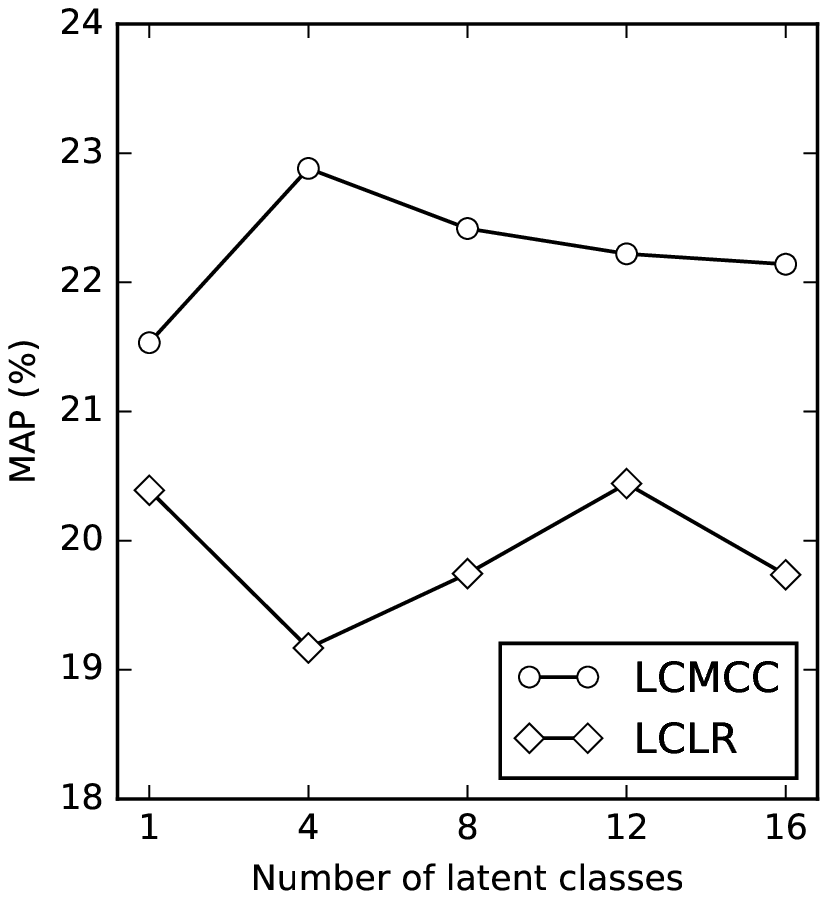}
&\includegraphics[clip,width=150pt]{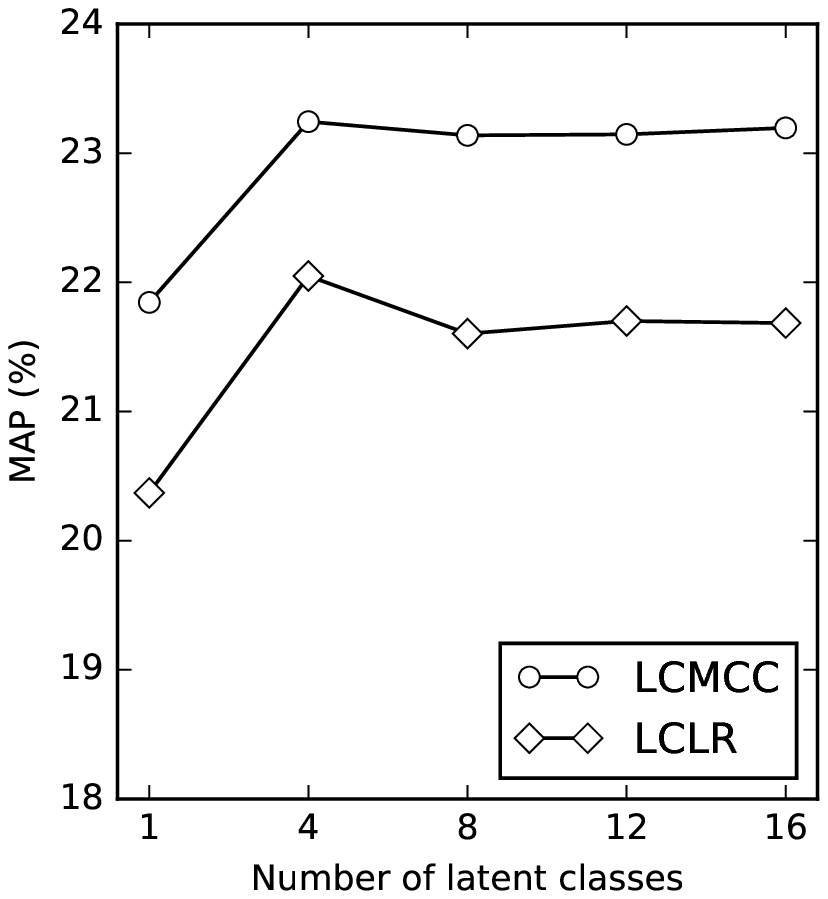}	&\includegraphics[clip,width=150pt]{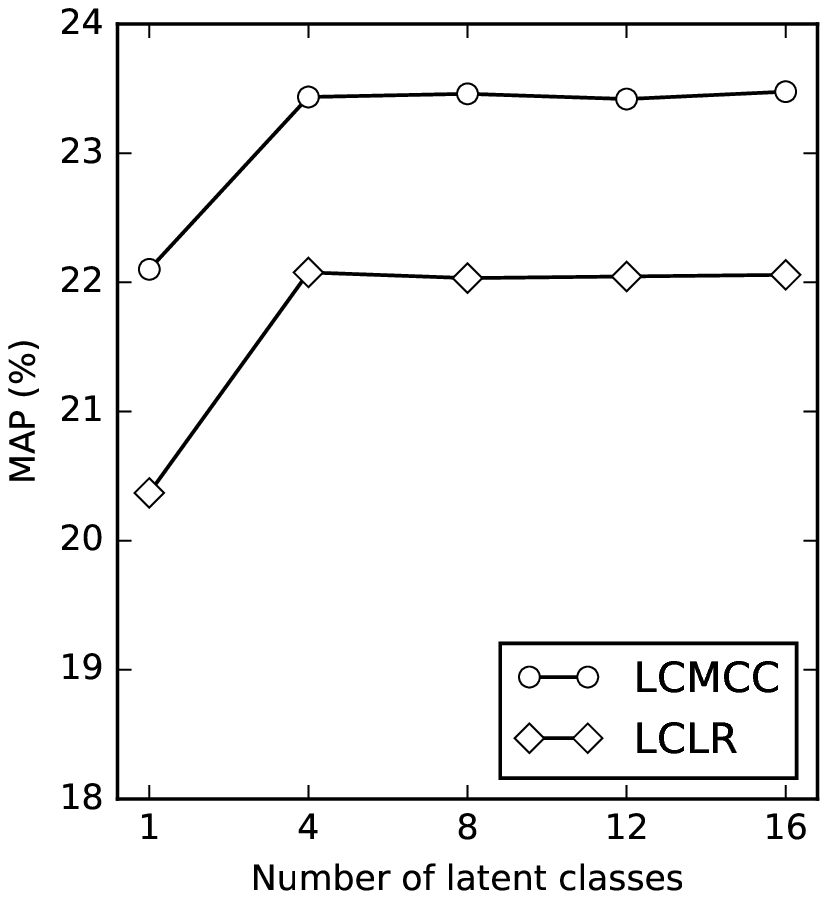}	\\ 
{\small (a) $1\%$-sampled}&{\small (b) $10\%$-sampled}&{\small (c) $100\%$-sampled}\\
\end{tabular}
\caption{Mean average precision of LCMCC and LCLR models for sampled training sets \label{fig:4}}
\end{figure}

Figure~\ref{fig:4} shows the mean average precisions (MAPs) of LCMCC($|S|$) and LCLR($|S|$) for the 1\%-, 10\%-, and 100\%-sampled sets, where the number of latent classes is $|S| \in \{1,4,8,12,16\}$. 
Here, MAP corresponds roughly to the average area under the precision-recall curve (see Manning et al.~\cite{MaRa08} for details) and represents the predictive performance that is independent of the number of selected products. 
Figure~\ref{fig:4} also confirms the superiority of our LCMCC model over the LCLR model in relation to predictive performance. 

Table~\ref{tbl:comp_overall} shows the computation time required by the EM algorithms to estimate the model parameters of LCMCC($|S|$) and LCLR($|S|$) for the 1\%-, 10\%-, and 100\%-sampled sets, where the number of latent classes is $|S| \in \{1,4,8,12,16\}$. 
We reiterate here that the EM-algorithm process was repeated 10 times with random initial estimates; it is the total computation time that is shown in Table~\ref{tbl:comp_overall}. 
Note also that the ``Preprocessing'' column gives the times for computing $(n_{ijk},q_{ijk})$ for $(i,j,k) \in I \times J \times K$ from the clickstream data, and that both LCMCC($|S|$) and LCLR($|S|$) require the same length of preprocessing time. 

We can see from Table~\ref{tbl:comp_overall} that the computation time increased with the number of latent classes. 
Meanwhile, there was little difference in computation time between LCMCC($|S|$) and LCLR($|S|$) for each $|S| \in \{1,4,8,12,16\}$. 

\begin{table}[tb]
\centering
\caption{Computation times of LCMCC and LCLR models in seconds \label{tbl:comp_overall}} 
\begin{tabular}{lrrrrrrrrrrr}
\toprule
Training set		& $|S|$	&	Preprocessing	&LCMCC		&LCLR		\\	\midrule
1$\%$-sampled 		&	1	&	4.0				&45.9		&39.8	\\	
					&	4	&					&1,626.3	&2,203.8	\\	
					&	8	&					&4,842.8	&5,227.7	\\	
					&	12	&					&9,626.5	&9,599.7	\\	
					&	16	&					&16,090.4	&14,431.6	\\	\midrule
10$\%$-sampled 		&	1	&	33.8			&46.6		&46.0		\\	
					&	4	&					&2,413.3	&2,771.6	\\
					&	8	&					&5,273.1	&6,463.1	\\	
					&	12	&					&10,343.9	&11,418.0	\\	
					&	16	&					&16,913.9	&15,899.2	\\	\midrule
100$\%$-sampled		&	1	&	338.6			&55.8		&65.5		\\	
					&	4	&					&2,100.0	&2,953.7	\\	
					&	8	&					&6,006.1	&6,463.9	\\	
					&	12	&					&11,206.1	&11,409.7	\\	
					&	16	&					&17,701.4	&15,741.7	\\	\bottomrule
\end{tabular}
\end{table}

\subsection{Analysis of latent classes of product categories}\label{sec:aelc}

We analyze the latent classes of product categories for the original (100$\%$-sampled) training set. 
Table~\ref{tbl:summary} summarizes the latent classes created by LCMCC(4). 
Here, product category $k$ is assigned to latent class $s$ that has the highest estimate of membership degree. 
Note also that in Table~\ref{tbl:summary}, product categories are listed in descending order of number of PVs in each latent class. 

\begin{table}[tb]
\centering
\caption{Summary on latent classes created by LCMCC(4)\label{tbl:summary} }
\begin{tabular}{ccl} \toprule
Class	&$\pi_s$	&Product categories											\\ \midrule
$s = 1$	&0.268		&Ladies' clothing, Shoes, Sports, Bags, Accessories, Furniture,	\\
		&			&Outdoors, Kids, Men's clothing, Jewelry, Golf,				\\
		&			&Storage furniture, Bicycles, Watches, Interior.				\\ \midrule
$s = 2$	&0.161		&Food, Sweets, Soft drinks/Coffee/Tea, Cosmetics, 			\\
		&			&PC accessories, Health, Diets, CDs/DVDs, Cat supplies/toys.		\\ \midrule
$s = 3$	&0.232		&Water, Books, Rice/Cereal, Contact lenses, Beer, Wine, Comics,	\\
		&			&Supplements, Medicine, Dog food, Cat food, Liqueurs, Spirits.	\\ \midrule
$s = 4$	&0.339		&Everyday sundries, Phone accessories, Home electronics,		\\
		&			&Underwear, Stationery, Baby, Toys, Kitchen supplies,			\\
		&			&Car supplies, Bedding, Dishes, Dog supplies/toys, DIY, 		\\
		&			&Handicrafts, Flowers, Gifts, Gardening, Games.					\\ \bottomrule
\end{tabular}
\end{table}

\begin{figure}[tb]
\centering
\begin{tabular}{cccc}
\includegraphics[clip,width=200pt]{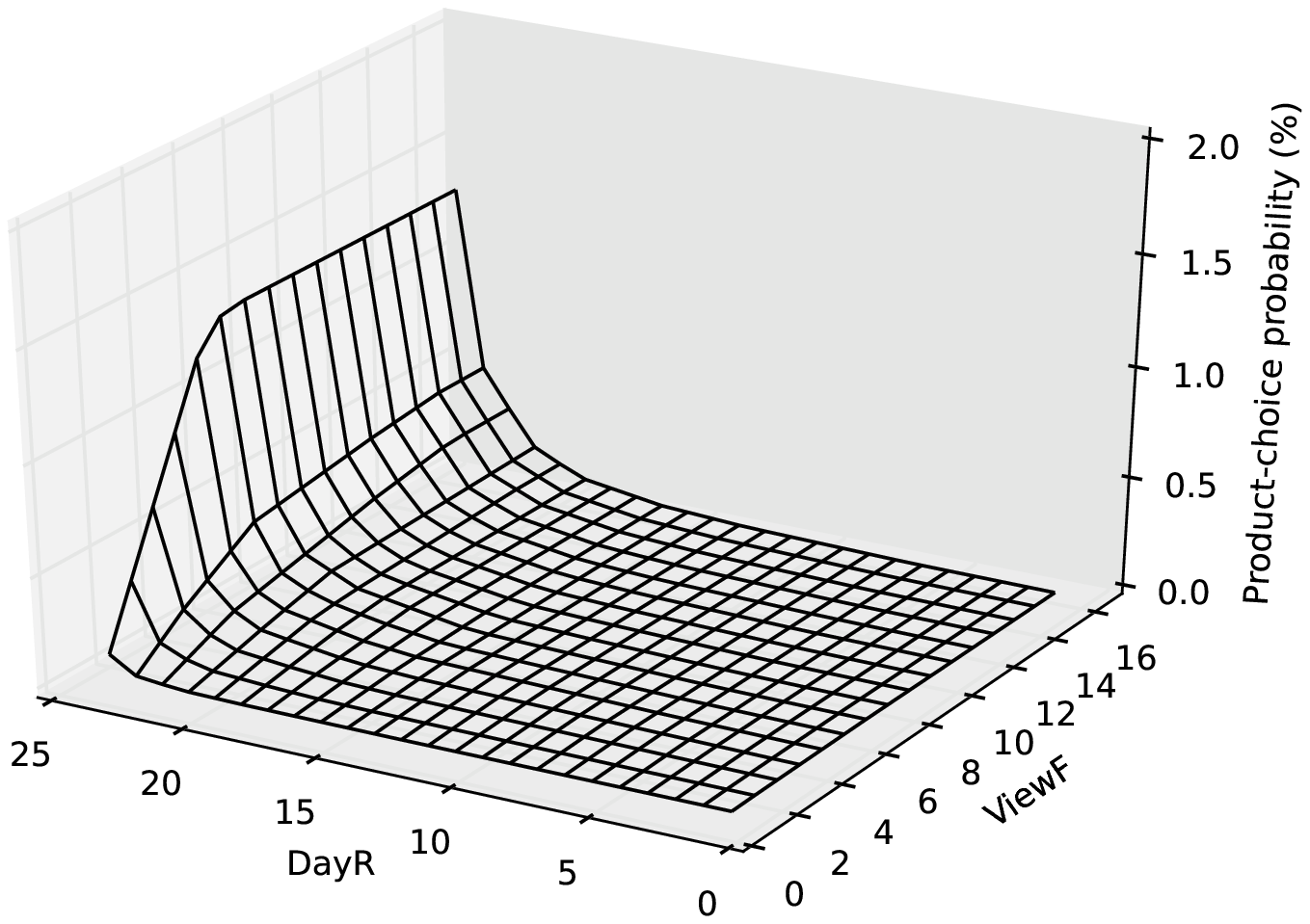} & \includegraphics[clip,width=200pt]{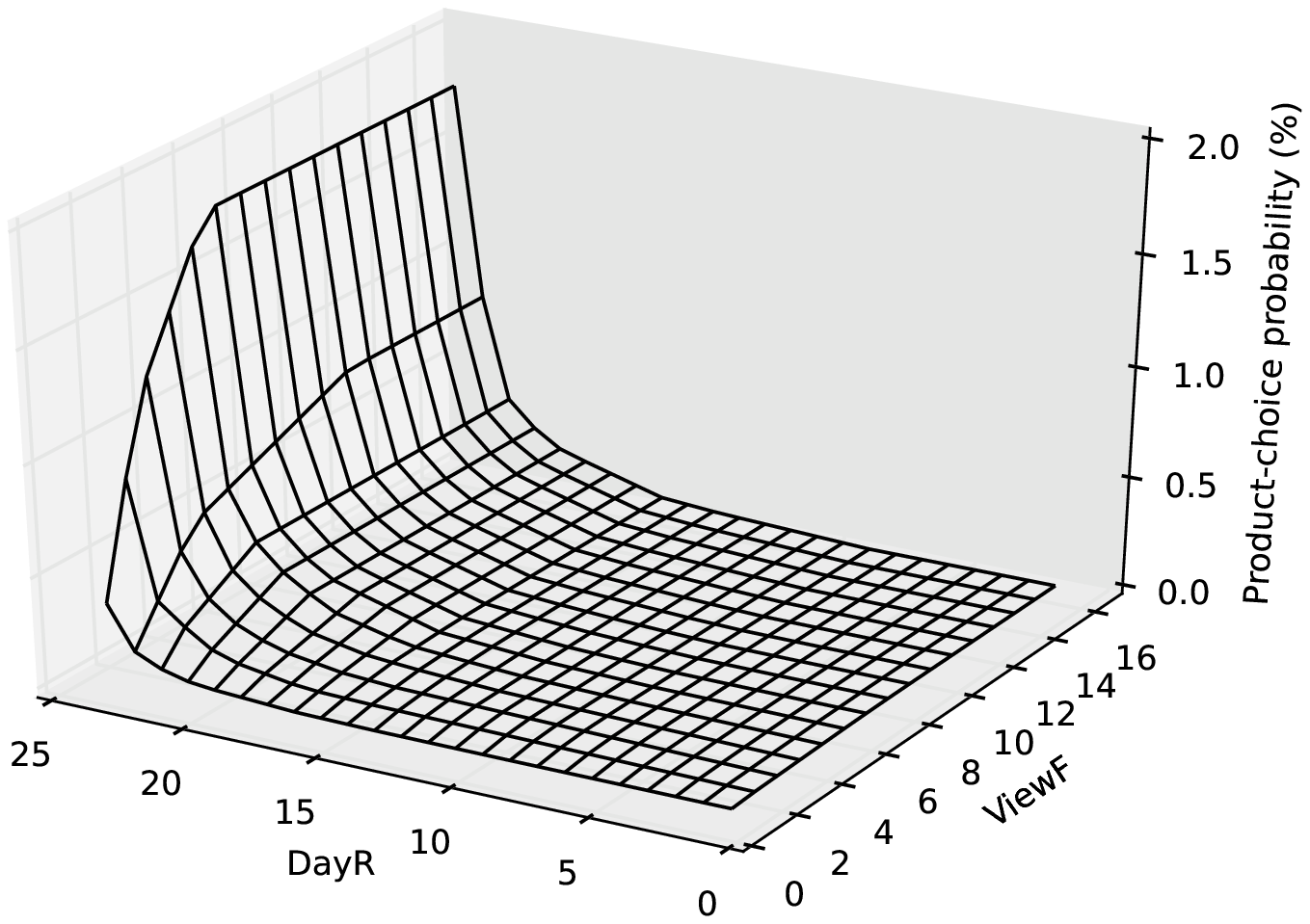}\\
{\small (a) Class 1} & {\small (b) Class 2} \\
\includegraphics[clip,width=200pt]{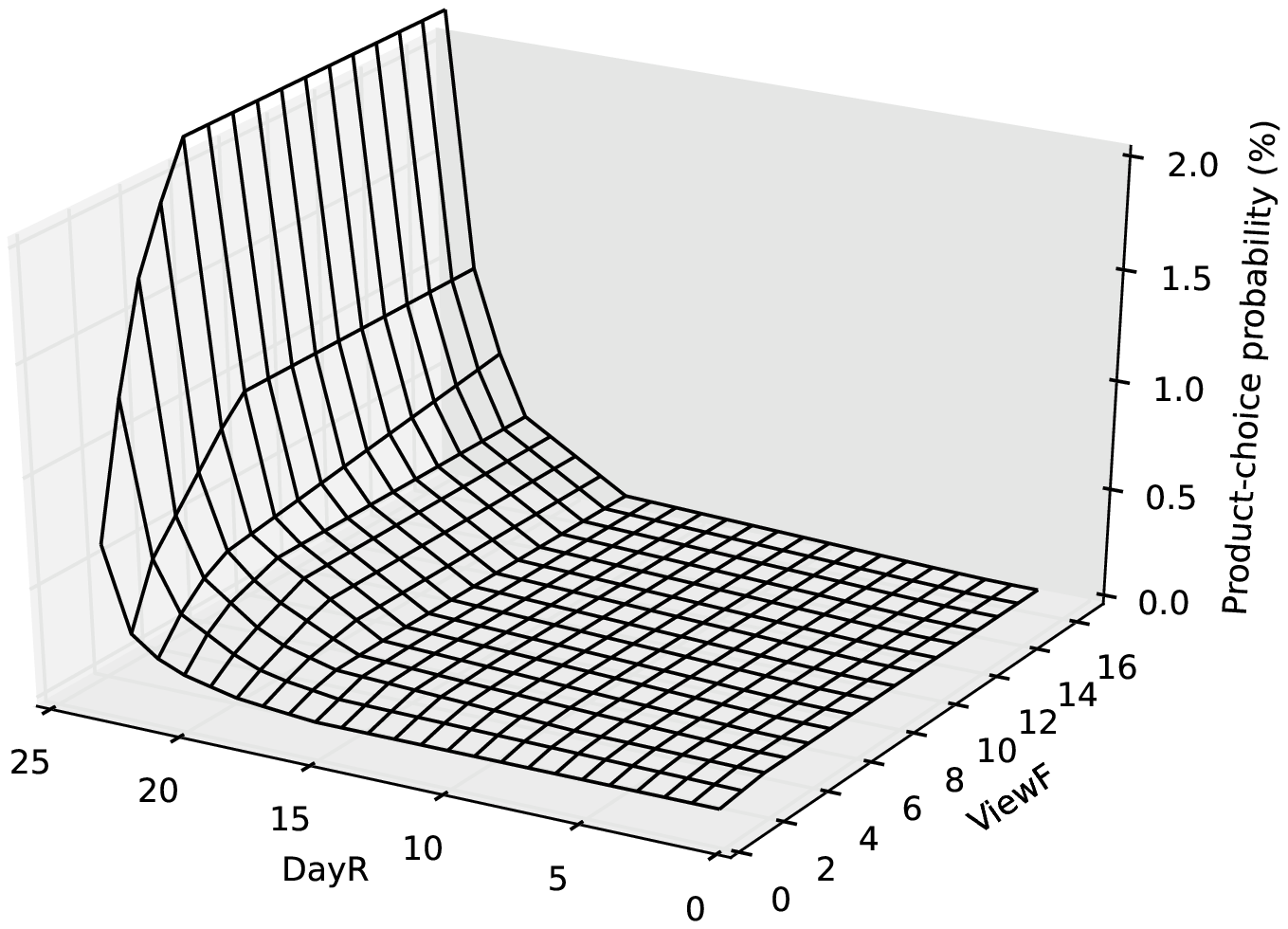} & \includegraphics[clip,width=200pt]{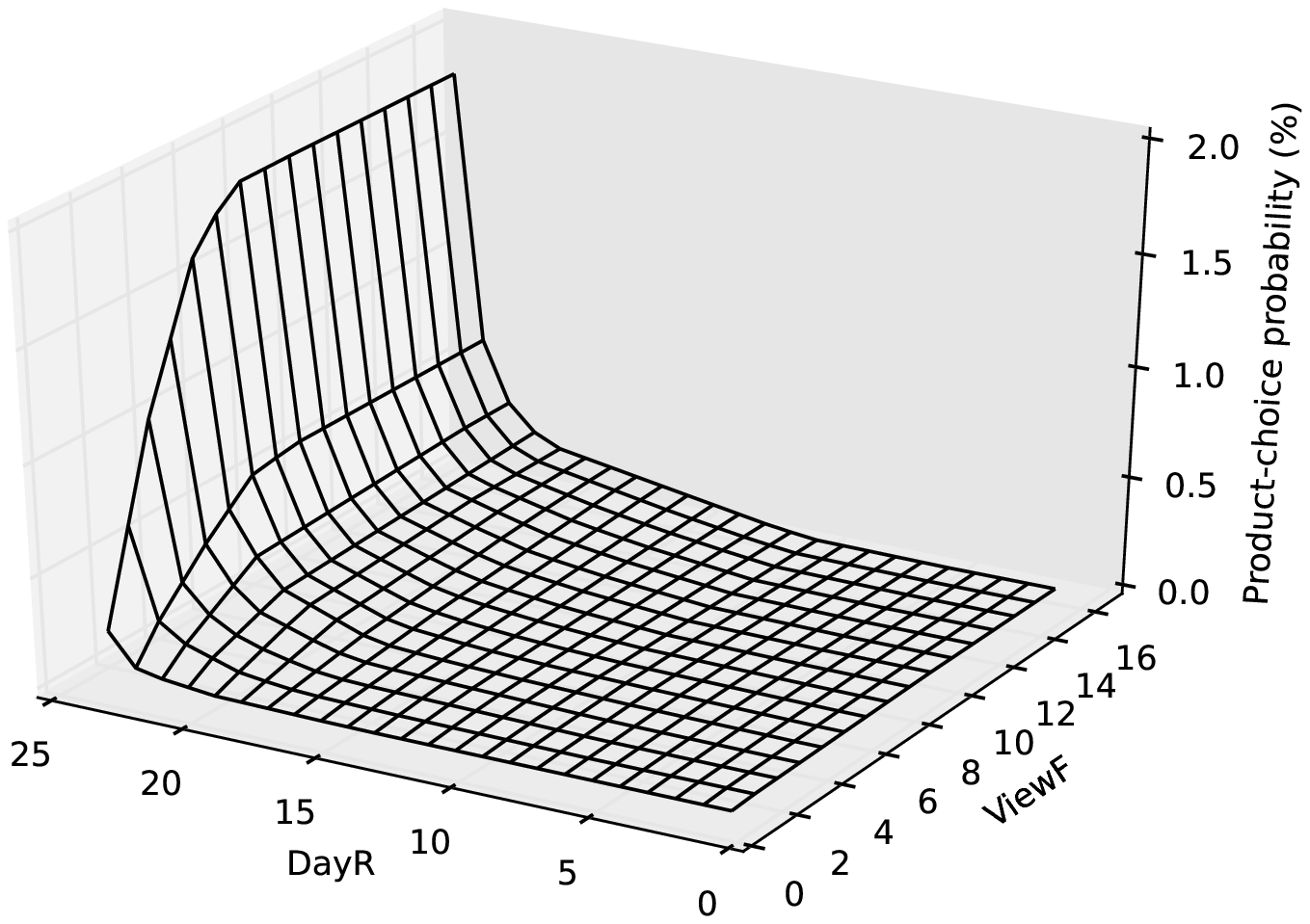}\\
{\small (c) Class 3} & {\small (d) Class 4} \\ 
\end{tabular}
\caption{Two-dimensional probability tables estimated by LCMCC(4)\label{fig:5}}
\end{figure}

Figure~\ref{fig:5} shows the product-choice probabilities estimated by LCMCC(4) for each latent class of the product categories. 
These product-choice probabilities are represented as a two-dimensional non-decreasing function with respect to recency and frequency because of the monotonicity constraints. 
They are also convex in recency value and concave in frequency value because of the convexity--concavity constraints. 
We observe that the product-choice probability rises sharply with recency value. 
The frequency also drives up the product-choice probability, especially when the recency value is very high. 

\begin{figure}[tb]
\centering
\begin{tabular}{cccc}
\includegraphics[clip,width=200pt]{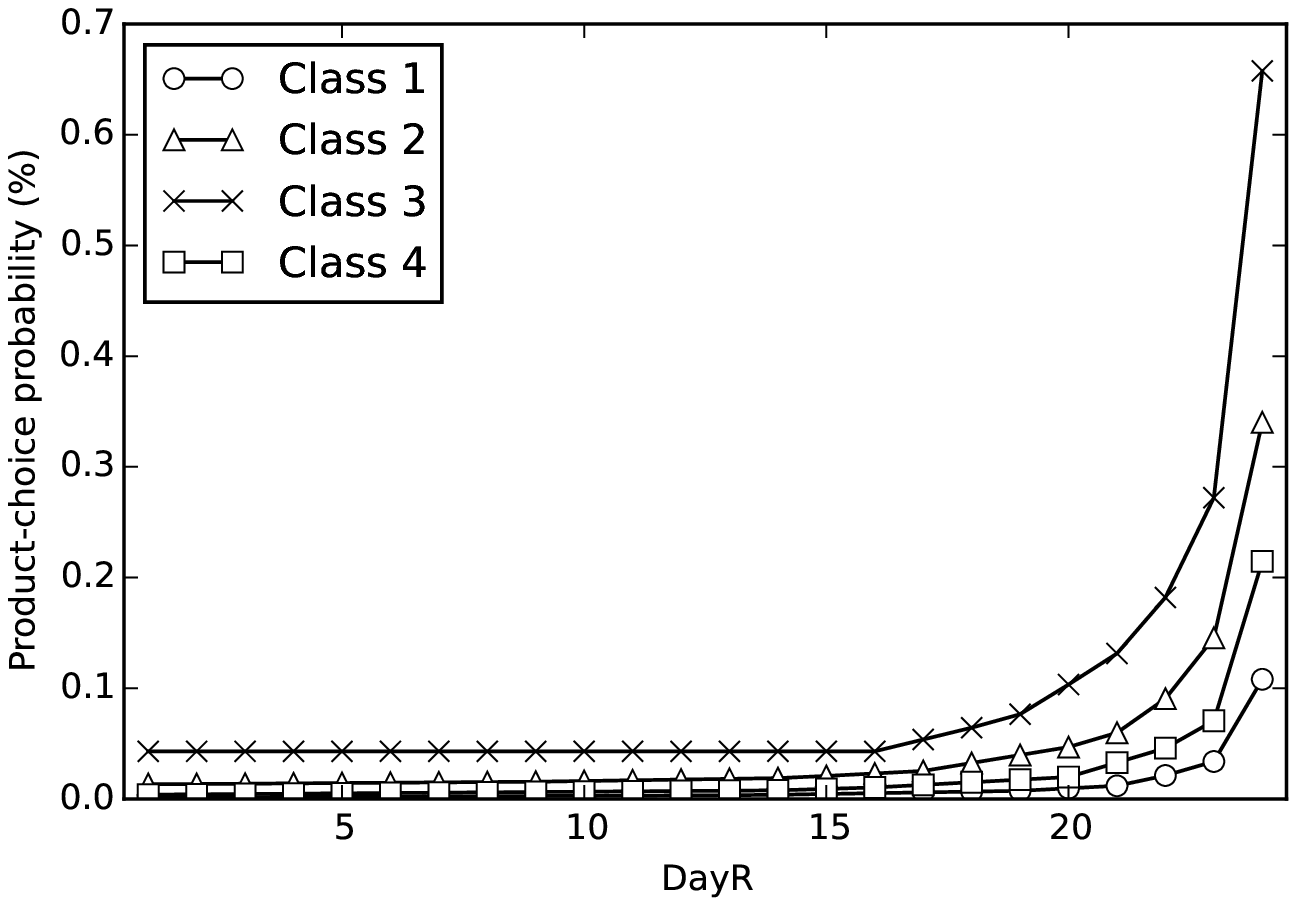}	&&\includegraphics[clip,width=200pt]{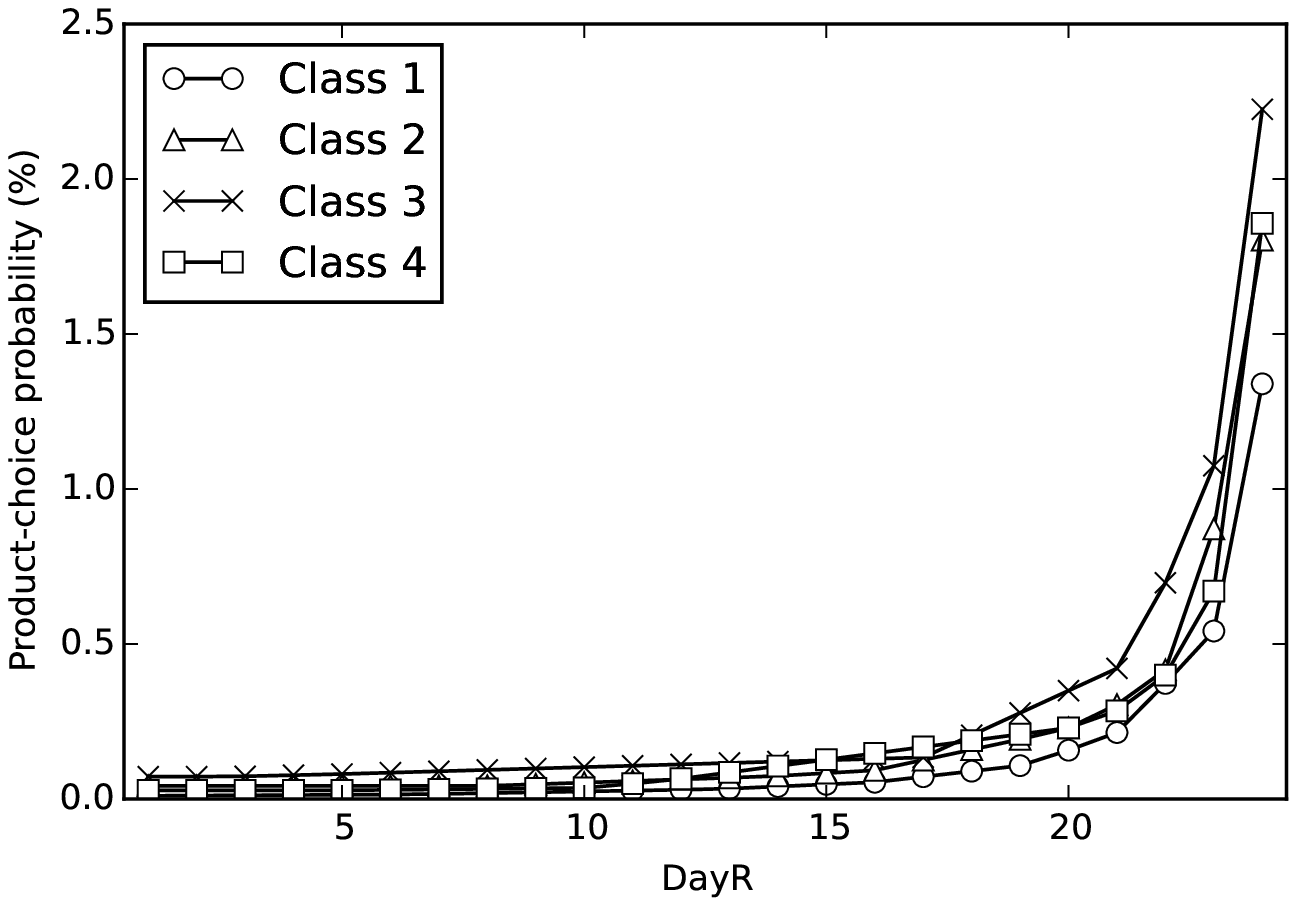}\\
{\small (a) ${\rm ViewF} = 1$}&&{\small (b) ${\rm ViewF} = 16$} \\ 
\includegraphics[clip,width=200pt]{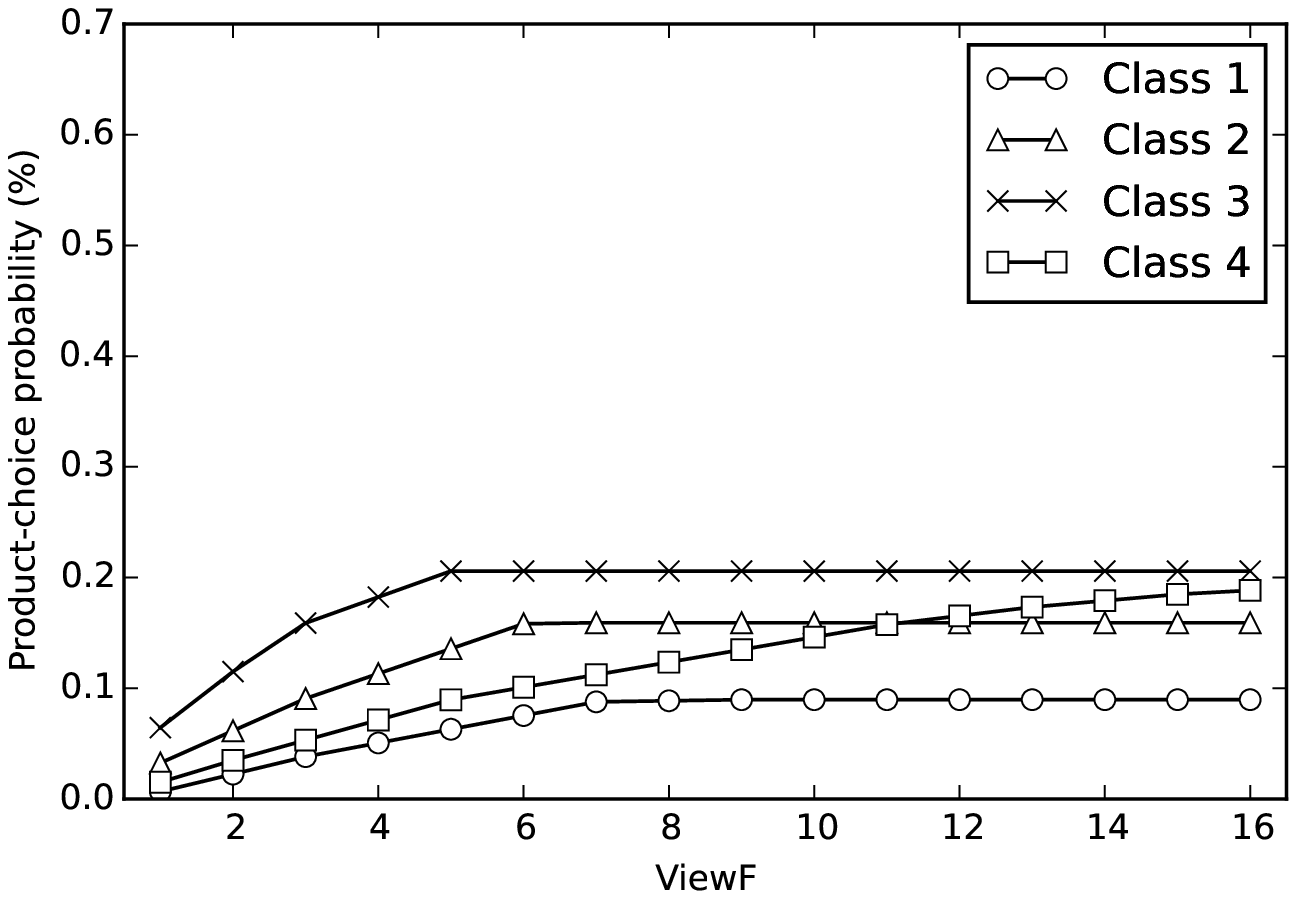}	&&\includegraphics[clip,width=200pt]{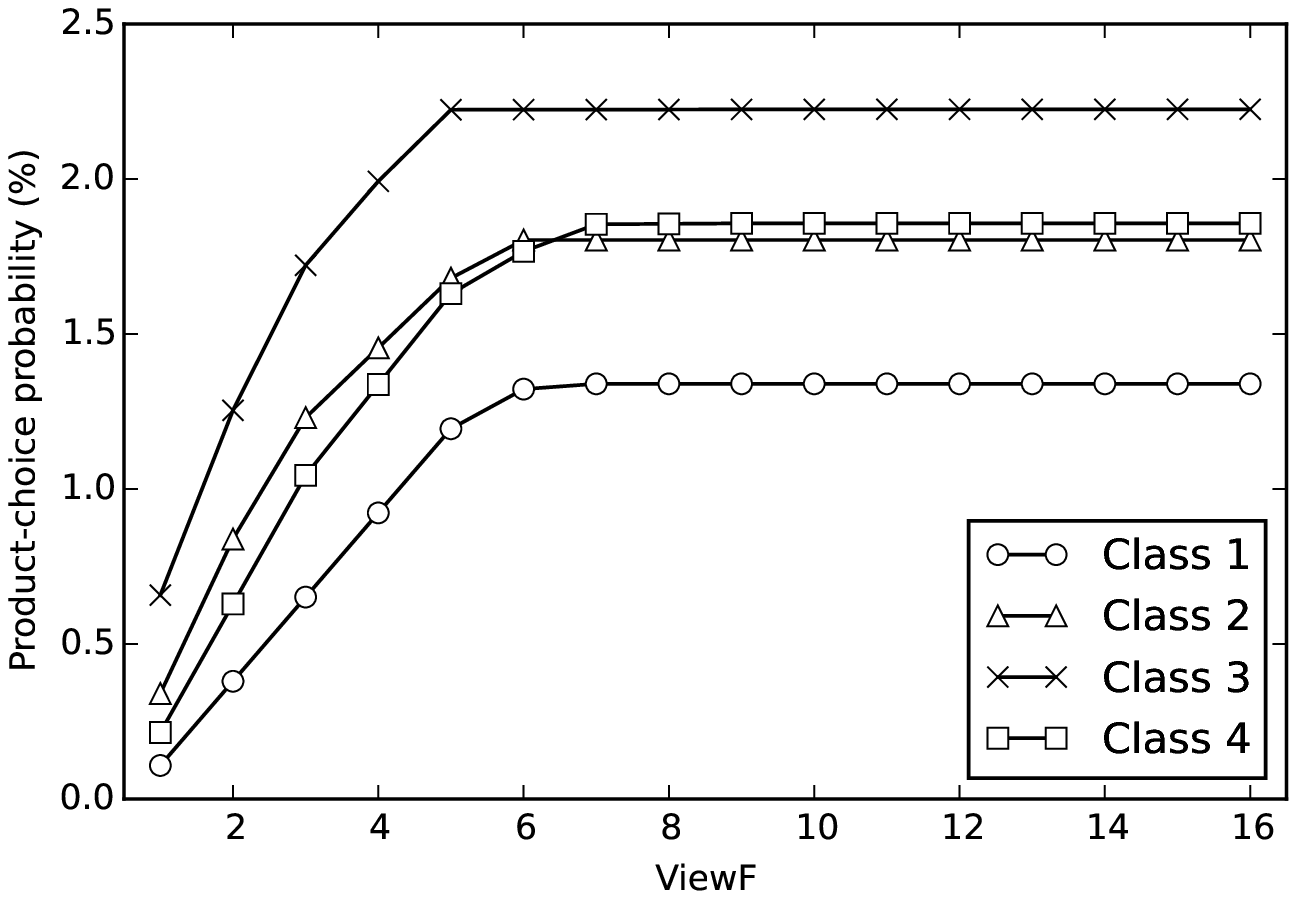}\\
{\small (c) ${\rm DayR} = 18$}&&{\small (d) ${\rm DayR} = 24$} \\ 
\end{tabular}
\caption{Product-choice probabilities estimated by LCMCC(4) for fixed recency or frequency value\label{fig:6}}
\end{figure}

To highlight the differences between these four probability tables, we express the product-choice probability as a function of one variable by fixing either the recency or the frequency to a certain value, as shown in Figure~\ref{fig:6}. 
In view of Table~\ref{tbl:summary} and Figures~\ref{fig:5} and \ref{fig:6}, we identify the characteristics of the four latent classes of product categories as follows. 

\paragraph{Class 1}
This class is composed mainly of clothing items and furnishings, as shown in Table~\ref{tbl:summary}. 
As we can see from Figures~\ref{fig:5} and \ref{fig:6}, the product-choice probabilities of class 1 are lower than those of the other classes. 
This is probably because customers of the e-commerce site repeatedly browse and compare various products in this class, such as ``Ladies' clothing,'' ``Shoes,'' ``Bags,'' and ``Accessories.'' 
In addition, many customers are likely to inspect and purchase such products at a brick-and-mortar store after having seen them on e-commerce sites. 

\paragraph{Class 2}
This class contains product categories such as ``Food,'' ``Soft drinks/Coffee/Tea,'' ``Cosmetics,'' and ``PC accessories.'' 
Because these products are relatively inexpensive, they are likely to be purchased even if their frequency value is not very high. 
Indeed, the product-choice probabilities of class 2 are the second highest when ${\rm ViewF} = 1$, as shown in Figure~\ref{fig:6}(a). 
Moreover, in Figure~\ref{fig:6}(c) and (d), the product-choice probabilities of class 2 are higher than those of class 4 only when ${\rm ViewF}$ is less than a certain value.  

\paragraph{Class 3}
This class consists mainly of products for daily use, such as ``Water,'' ``Rice/Cereal,'' ``Contact lenses,'' and ``Dog food.'' 
Figure~\ref{fig:6} shows that the product-choice probabilities of class 3 are always the highest of the four classes. 
This implies that customers regularly purchase favorite products in this class without viewing the product details repeatedly. 

\paragraph{Class 4}
This class is composed mainly of more expensive products such as ``Home electronics,'' ``Stationery,'' ``Kitchen supplies,'' and ``Car supplies.'' 
We note also that ``Everyday sundries'' contains durable and expensive goods such as electric toothbrushes, suitcases, and high-quality towels. 
Since these products are not purchased frequently, customers may purchase them after checking the product details repeatedly. 
Indeed, Figure~\ref{fig:6}(c) reveals that the product-choice probability of class 4 increases monotonically with ViewF and exceeds that of class 2 at ${\rm ViewF} = 12$. 
In other words, products in class 4 become more likely to be purchased as the number of PVs increases. 

\section{Conclusions}

We devised a novel latent-class shape-restricted model for estimating product-choice probabilities from the recency and frequency of each customer's previous PVs. 
Our model classifies products into a specified number of latent classes, for each of which it estimates a two-dimensional probability table that satisfies monotonicity, convexity, and concavity constraints with respect to recency and frequency. 
We also developed a specialized EM algorithm for estimating the parameters of our model. 
Moreover, we analyzed customer product-choice behavior based on actual clickstream data by means of our latent-class model. 

Our research contribution is to open up a new possibility for latent-class modeling in the analysis of clickstream data. 
Specifically, we successfully improved the predictive performance of the existing MCC model by applying latent-class modeling. 
We also established a new method of product cluster analysis based on the recency and frequency of PVs on e-commerce sites. 

The computational results demonstrated clearly that our latent-class MCC model outperformed common latent-class logistic regression. 
Since this fact proves the effectiveness of combining latent-class regression with shape-restricted regression, this paper will motivate further work on latent-class shape-restricted regression in the areas of marketing and business research. 

The estimated product-choice probabilities represent a customer's preferences for the products. 
Such information is essential in recommender systems (e.g., collaborative filtering), and thus, the performance of recommender systems on e-commerce sites would be enhanced by incorporating our probability tables. 
Additionally, once probability tables have been developed by the EM algorithm, they can easily be put into practical use on e-commerce sites. 
Furthermore, our latent-class model provides useful information for formulating an individual sales promotion strategy for each product category. 

A future direction of study is the effective use of latent classes of customers in the shape-restricted model. 
Our preliminary experiments show that the use of latent customer classes does not lead to improved predictive performance on the clickstream data analyzed in this paper. 
However, since the modeling of customer heterogeneity is the central challenge of many statistical marketing studies, the use of latent customer classes will offer certain advantages in other application areas. 

\section*{Acknowledgments}
The authors would like to thank Recruit Lifestyle Co., Ltd. for providing the clickstream data used in the computational experiments. 





\end{document}